%% file: example_paper.tex
\documentclass{article}

\usepackage{microtype}
\usepackage{graphicx}
\usepackage{subcaption}
\usepackage{booktabs} 
\usepackage{enumitem}
\usepackage[table]{xcolor}
\usepackage{multirow}
\usepackage{tcolorbox}
\tcbuselibrary{breakable}
\usepackage{fontawesome}
\usepackage{hyperref}

\newcommand{\maze}{\textsc{MazeNavigation}}
\newcommand{\tangram}{\textsc{TangramPuzzle}}

\newcommand{\rparagraph}[1]{\noindent\textbf{#1}~~}

\newcommand{\rparagraphnospace}[1]{\noindent\textbf{#1}}

\usepackage[preprint]{icml2026}

\usepackage{amsmath}
\usepackage{amssymb}
\usepackage{mathtools}
\usepackage{amsthm}

\usepackage[capitalize,noabbrev]{cleveref}

\theoremstyle{plain}

\theoremstyle{definition}

\theoremstyle{remark}

\usepackage[textsize=tiny]{todonotes}

\icmltitlerunning{Thinking in Frames: How Visual Context and Test-Time Scaling Empower Video Reasoning}

\begin{document}

\twocolumn[
  \icmltitle{Thinking in Frames: How Visual Context and\\ Test-Time Scaling Empower Video Reasoning}
  \icmlsetsymbol{equal}{*}

  \begin{icmlauthorlist}
    \icmlauthor{Chengzu Li}{equal,cam,ku}
    \icmlauthor{Zanyi Wang}{equal,ucsd}
    \icmlauthor{Jiaang Li}{equal,ku}
    \icmlauthor{Yi Xu}{cam}
    \icmlauthor{Han Zhou}{cam}
    \icmlauthor{Huanyu Zhang}{casia}  
    \icmlauthor{Ruichuan An}{pku}   \\
    \icmlauthor{Dengyang Jiang}{hkust}
    \icmlauthor{Zhaochong An}{ku}
    \icmlauthor{Ivan Vulić}{cam}
    \icmlauthor{Serge Belongie}{ku}
    \icmlauthor{Anna Korhonen}{cam}
  \end{icmlauthorlist}

  \icmlaffiliation{cam}{University of Cambridge}
  \icmlaffiliation{ku}{Pioneer Center for AI, University of Copenhagen}
  \icmlaffiliation{hkust}{Hong Kong University of Science and Technology}
  \icmlaffiliation{ucsd}{University of California San Diego}
  \icmlaffiliation{casia}{Institute of Automation, Chinese Academy of Sciences}
  \icmlaffiliation{pku}{Peking University}

  \icmlcorrespondingauthor{Chengzu Li}{cl917@cam.ac.uk}
  \icmlkeywords{Machine Learning, ICML}

  \vskip 0.3in
]

\printAffiliationsAndNotice{\icmlEqualContribution}

\input{secs/abstract}
\input{secs/1_introduction}

\input{secs/2_related_work}
\input{secs/3_method}

\input{secs/4_experiments}
\input{secs/5_discussion}
\input{secs/6_conclusion}

\input{secs/impact_statement}

\bibliography{example_paper}
\bibliographystyle{icml2026}

\newpage
\appendix
\onecolumn
\input{appendix/A_experiment_setups}
\input{appendix/B_more_results}

\input{appendix/C_prompts}

\end{document}

%% file: secs/abstract.tex
\begin{abstract}
    Vision-Language Models have excelled at textual reasoning, but they often struggle with fine-grained spatial understanding and continuous action planning, failing to simulate the dynamics required for complex visual reasoning. 
    In this work, we formulate visual reasoning by means of video generation models, positing that generated frames can act as intermediate reasoning steps between initial states and solutions. 
    We evaluate their capacity in two distinct regimes: \maze~for sequential discrete planning with low visual change and \tangram~for continuous manipulation with high visual change. 
    Our experiments reveal three critical insights:
    \textit{(1) Robust Zero-Shot Generalization}: 
    In both tasks, the model demonstrates strong performance on unseen data distributions without specific finetuning.
    \textit{(2) Visual Context}: 
    The model effectively uses visual context as explicit control, such as agent icons and tangram shapes, enabling it to maintain high visual consistency and adapt its planning capability robustly to unseen patterns.
    \textit{(3) Visual Test-Time Scaling}: We observe a test-time scaling law in sequential planning; increasing the generated video length (visual inference budget) empowers better zero-shot generalization to spatially and temporally complex paths. 
    These findings suggest that video generation is not merely a media tool, but a scalable, generalizable paradigm for visual reasoning.
    \vspace{-2mm}
\end{abstract}

%% file: secs/1_introduction.tex
\section{Introduction}

The ability for reasoning and planning, decomposing a complex goal into actionable steps, has been approached through Large Language Models (LLMs) \citep{achiam2023gpt,yang2025qwen3,comanici2025gemini}.
However, while Multimodal Large Language Models (MLLMs) have achieved remarkable success in understanding and reasoning over visual semantics \citep{Bai2025Qwen3VLTR,bai2025qwen2}, they face significant limitations in embodied and spatial domains for reasoning \citep{li-etal-2024-topviewrs,yang2025embodiedbench}. 
One of the bottlenecks lies in the medium of reasoning modality: MLLMs typically reason via textual descriptions, which are often inefficient and imprecise for capturing fine-grained geometric contexts and physical dynamics \citep{gao2025gllava,li202511plus,fu2025geolaux} (e.g., describing the exact continuous rotation and placement of a tangram piece or a collision-free trajectory in an irregular maze). 

Recognizing these limitations, recent research has shifted toward reasoning with multimodal traces, such as MVoT \citep{li2025imagine} and Visual Planning \citep{xu2025visual}, proposing that thinking directly in the visual domain provides a higher-bandwidth representation for spatial intelligence.
Concurrently, video generation models have been viewed primarily as tools for media creation, optimized for aesthetic quality rather than logical consistency \citep{huang2024vbench,wan2025wan}.
While recent proprietary models like Veo 3 \citep{google_veo3} have qualitatively explored using video for reasoning, these efforts lack in-depth quantitative investigation and open reproducibility \citep{wiedemer2025video}.
We argue that video generation offers a unique paradigm for reasoning: by operating on a dense temporal manifold, the model acts as a continuous proxy of the reasoning process. 
This offers better interpretability and verifiability compared to single-step image editing \citep{cai2025z,tong2026cof}, as it captures the process of change.

\input{figures/main_fig}

Despite the promise of visual expressiveness, employing generative video for planning introduces inherent challenges regarding controllability and fidelity. 
Prior work in visual reasoning \citep{xu2025visual,guo2025video,tong2025thinking} has largely focused on tasks with discrete actions such as maze navigation, where visual changes between frames are minimal. 
While these studies establish a foundation, they have been confined to in-distribution settings, for instance, evaluating models on familiar grid sizes and patterns for maze navigation \citep{xu2025visual,he2025diffthinker}. 
A key aspect of reasoning, however, also involves \textit{generalization}~\citep{huang2025mathperturb}: the ability to apply learned rules, such as collision avoidance or motion control, to unseen difficulty levels (e.g., larger mazes) and novel visual contexts (e.g., unseen agent avatars or object shapes).
Furthermore, prior work doesn't fully evaluate models' capabilities in controllability and fidelity as generative visual reasoners. 
Maintaining complex geometric consistency under continuous action spaces with high visual changes remains challenging for generative models \citep{wang2025deforming,wu2025hunyuanvideo}, and existing studies obscure these difficulties by focusing on simpler or highly constrained settings.

In this work, we conduct an in-depth investigation of video generation models for visual planning, aiming to understand whether and how these models support planning under diverse challenges.
To cover the spectrum of visual reasoning, we evaluate two distinct regimes: (1) \maze~representing sequential, structured planning with low visual change and (2) \tangram~for continuous manipulation with high visual change. 
Unlike prior work, \maze~is investigated with various out-of-distribution settings, and \tangram~forces the model to handle dynamic visuals, requiring precise geometric preservation during continuous manipulation.

Our experiments reveal that video-based reasoning consistently outperforms text-based baselines, particularly in \tangram~where MLLMs struggle to verbalize precise spatial manipulations. 
Beyond raw performance, we observe two critical emergent properties.
First, the model demonstrates \textit{Visual Context as Control}: by conditioning on specific visual context such as agent icons or puzzle shapes, the model maintains high visual consistency and robustly adapts to unseen patterns without fine-tuning. 
This allows for strong zero-shot performance across spatially and visually Out-Of-Distribution (OOD) settings.
Second, we uncover a \textit{Visual Test-Time Scaling Law} in sequential planning that parallels ``System 2'' thinking in human cognition \citep{kahneman2011thinking}.
We observe that increasing the generated video frames, which effectively increases the visual inference budget, empowers the model to solve spatially and temporally complex paths that fail at lower frame counts (e.g., extending the video length from 81 to 121 frames).
We posit that in the visual domain, temporal frames can also act as a compute budget for reasoning, allowing the model to ``think'' for longer about the solution.

The main contributions of this paper include:
\begin{itemize}[leftmargin=*,nosep]
    \item \textit{New Regime for Visual Reasoning:~} In addition to OOD situations for \maze, we introduce \tangram, which shifts spatial reasoning from static, discrete planning to a continuous dynamic process capable of handling high visual change with geometric constraints.
    \item \textit{Visual Context for Robust Generalization:~} We demonstrate that video generation models can abstract underlying planning algorithms rather than memorizing pixels with OOD visuals and maintain geometric consistency by offering visual context as control. 
    \item \textit{Visual Test-Time Scaling:~} We identify a scaling law in sequential planning, showing that increasing the inference frame budget acts as test-time compute, significantly improving generalization on long-horizon, complex tasks.
\end{itemize}

%% file: figures/main_fig.tex
\begin{figure*}[t
]
    \centering
    \includegraphics[width=\linewidth]{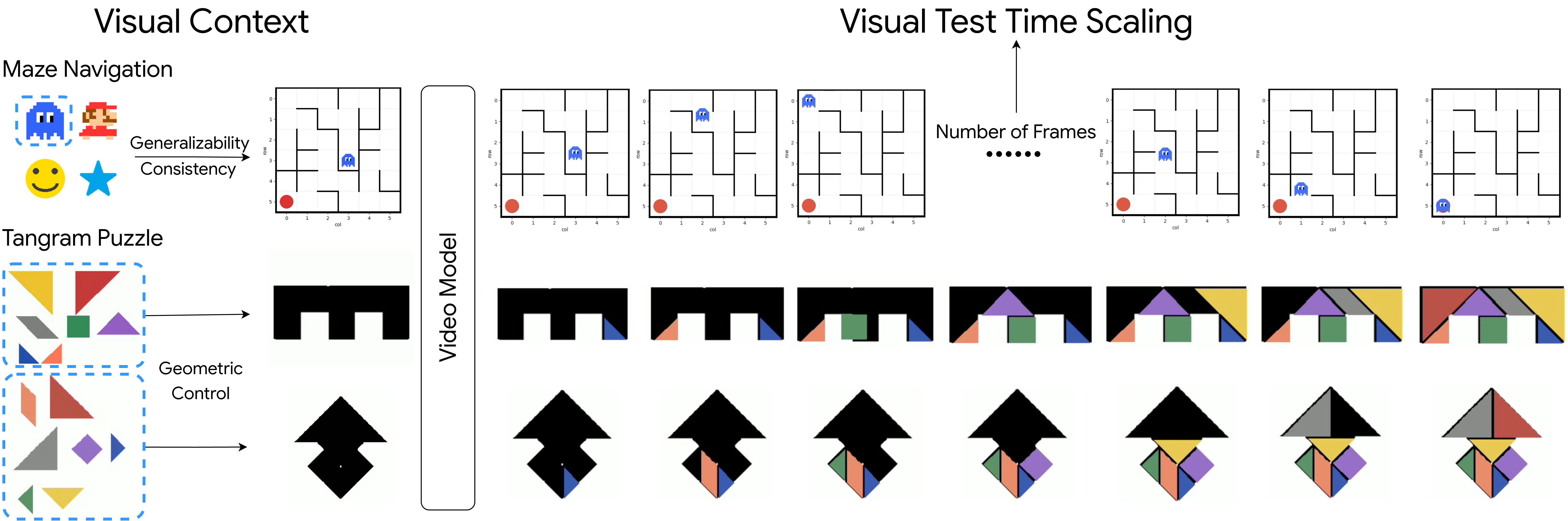}
    \caption{Video generation models as visual reasoners, empowered by (1) enriched visual context for improved geometric control and (2) visual test-time scaling that allocates a larger inference-frame budget and enables stronger performance on long-horizon, complex sequential planning tasks, together demonstrating robust generalization across diverse scenarios.}
    \label{fig:main fig}
\end{figure*}

%% file: secs/2_related_work.tex
\section{Related Work}

\rparagraph{Multimodal Planning.}
To extend textual reasoning traces to multimodality,
one line of research employs MLLMs and grounds the reasoning process through interleaved text-image sequences generated by external tools. 
These tools include symbolic programming languages used as sketchpads \citep{hu2024visual}, explicit coordinate-based representations to enhance perceptual grounding \citep{su2025pixel,fan2025grit}, and other generative models invoked via tool calls \citep{zheng2025deepeyes,cheng2026omni}.
Beyond tool-based approaches, another line of work focuses on native multimodal visual planning by designing MLLM architectures capable of directly generating multimodal content. This includes image sequence \citep{xu2025visual} and interleaved multimodal traces generation \citep{li2025imagine}, as well as recursive mixed-modality generation \citep{li2025zebra,gu2025thinkmorph}.
A separate direction leverages latent representations for reasoning; however, these methods generally lack explicit visualized ``visual thoughts'' \citep{li2025latent,tong2025sketch} except for \citet{zhang2025latent}.
Despite these advances, prior work has largely focused on in-distribution settings and structured planning tasks, with limited investigation into out-of-distribution (OOD) generalization.
In addition to interleaved multimodal traces, recent studies have also explored video-based representations for visual planning, which we discuss in the following paragraph.

\rparagraph{Video Generation.}
Video generation models have traditionally been studied primarily as media creation tools, with an emphasis on visual fidelity, temporal coherence, and consistency as world models \citep{huang2024vbench,wan2025wan,team2025evaluating}.
More recently, efforts have begun to repurpose video generation models as visual reasoners, with qualitative evaluations conducted on both proprietary models \citep{wiedemer2025video} and open-source models, either in zero-shot settings \citep{guo2025video} or after fine-tuning \citep{yang2025reasoning}.
However, these studies largely focus on maze navigation, a form of sequential discrete action planning that involves minimal visual change across frames while preserving a fixed spatial layout.
Beyond their emphasis on in-distribution performance, the fundamental challenges faced by generative visual reasoners remain underexplored, particularly scenarios involving continuous manipulation with substantial visual changes over time, where maintaining geometric and spatial consistency throughout the video remains a significant challenge \citep{fan2024refdrop}.

\rparagraph{Test-Time Scaling.} 
Chain-of-Thought (CoT) prompting \citep{wei2022chain} enables models to solve complex problems by explicitly generating intermediate reasoning steps.
Recent work \citep{openai_reasoning_llms_2024} has shown that scaling test-time compute, i.e., increasing computation during inference either through multiple \textit{parallel} sampled inferences with aggregation \citep{brown2025large} or a single inference with longer traces \citep{muennighoff2025s1simpletesttimescaling}, can yield performance improvements analogous to scaling training data.
In the visual domain, however, such test-time scaling has not been systematically explored for video generation models.
\citet{Liu_2025_ICCV} investigated parallel test-time scaling for video generation, targeting higher perceptual quality rather than visual planning or reasoning.
In contrast, we are the first to identify a direct visual analogue of test-time scaling laws for planning. Specifically, we show that increasing the frame budget in video generation is empirically similar to increasing the length of reasoning traces in LLMs, enabling models to solve planning problems with OOD complexity.

%% file: secs/3_method.tex
\section{Video Generation for Visual Planning}

\subsection{Formulation}
Following the definition from \citet{xu2025visual}, we formally frame video generation as a visual planning problem.
Given an initial state image $s_{start} \in \mathcal{S}$, a goal specification $g$ (e.g., a target image or pattern), and a set of latent physical constraints $c$, the objective is to generate a video sequence $V = \{v_0, v_1, \dots, v_T\}$ where $v_0 = s_{start}$.
A valid solution trajectory must transition from $s_{start}$ to a final state $v_T$ that satisfies the goal condition $g$, while maintaining temporal consistency and adhering to constraints $c$ (e.g., collision avoidance, object permanence).
In this framework, the generative model $\mathcal{P}_\theta(V | s_{start}, g)$ acts as the planning policy, where the temporal evolution of frames $v_t \rightarrow v_{t+1}$ corresponds to the execution of the plan.
Unlike symbolic planners that output discrete actions $a_t$, the video model outputs continuous high-dimensional dense transitions, requiring the implicit learning of latent rules and causal dynamics.

\subsection{Spectrum of Visual Planning Regimes}
\label{subsec:regimes}
Prior research has predominantly focused on grid-world navigation (e.g., mazes) \citep{xu2025visual,guo2025video}.
While effective for testing sequential logic, these environments are \textit{visually static} with \textit{local visual change}: only a small agent moves against a fixed background.
Such tasks operate in \textit{discrete} action spaces that can often be solved via symbolic proxies \citep{dao2025alphamaze}, overlooking the unique advantage of generative video models: the ability to model complex visual dynamics in continuous action spaces.

To rigorously evaluate the model's capabilities, we define two contrasting reasoning regimes (summarized in Table \ref{tab:regimes}). 
In addition to \maze, \tangram~introduces a novel stress test for visual planning.
It requires the model to manage high visual change with geometric consistency, and continuous object manipulation, where text-based reasoning fundamentally fails. 

\rparagraph{\maze.}
We adopt standard maze navigation to maintain comparability with previous work \citep{xu2025visual,miniveo3reasoner}, where the agent icon moves along the maze to get the final destination. 
This task evaluates the model's ability to retain long-term consistency of the map structure and execute precise, collision-free pathfinding with minimal visual change of the agent along the path.

\rparagraph{\tangram.}
We introduce \tangram~as a novel challenge for continuous spatial manipulation. 
The objective is to manipulate a set of 7 disjoint geometric pieces from the tangram to precisely fill a target silhouette.
Unlike mazes, the difficulty for the visual generative model here is not path length but \textit{geometric preservation}, since the length is fixed by the number of pieces. 
Throughout the reasoning process, the entire scene changes as pieces are translated and rotated continuously.
Formally, given a target silhouette without any infilling pieces $s_{start}$, the model generates a solution $v_T$ (the final frame if using video generation) where all pieces $p_i$ are placed within $g$ without overlapping or distortion as constraints $c$.
This task presents a visual-native challenge where textual descriptions struggle to efficiently capture the precise relative locations and rotations required for a valid solution.

\input{tables/comparison_tasks}

%% file: tables/comparison_tasks.tex
\begin{table}[t]
\centering
\small
\caption{Comparison of Reasoning Regimes.}
\label{tab:regimes}
\resizebox{\linewidth}{!}{%
\begin{tabular}{lcc}
\toprule
\textbf{Aspect} 
& \textbf{\maze}
& \textbf{\tangram} \\
\midrule
Visual Volatility 
& Low 
& High \\
Action Space 
& Discrete 
& Continuous \\
Core Challenge 
& Long-horizon Logic 
& Spatial Geometry \\
\bottomrule
\end{tabular}
}
\end{table}

%% file: secs/4_experiments.tex
\section{Experiments}
We first describe the experimental setups (\S \ref{subsec:experimental_setups}) and analyze the performance of different systems on both tasks, considering not only in-distribution results (\S \ref{subsec:in-distribution results}) but also zero-shot generalization in out-of-distribution settings (\S \ref{subsec:ood results}).

\subsection{Experimental Setups}
\label{subsec:experimental_setups}
\rparagraph{Data.}
We construct the \maze~dataset following \citet{xu2025visual} and \citet{miniveo3reasoner}. 
The dataset covers varying levels of difficulty in both maze size (from $3\times3$ to $6\times6$) and visual appearance, including 40 distinct agent icons with different colors as shown in Figure \ref{appfig:maze icon} in Appendix \ref{appsubsec:dataset}. 
For each data, the optimal navigation path from the start location to the goal is determined using heuristic search algorithms \citep{ivanitskiy2023configurable}.
For \tangram, we build upon the dataset introduced by \citet{ji-etal-2022-abstract}, which evaluates whether MLLMs can recognize abstract patterns composed of tangram pieces. 
We repurpose this dataset into a sequential setting, where tangram pieces are placed into their target locations in the abstract pattern step by step. 
Additional details on data statistics, collection and preprocessing are provided in Appendix~\ref{appsubsec:dataset}. 

\rparagraph{Metrics.}
For \maze, we adopt the same metrics from \citet{xu2025visual} with Exact Match (EM) and Progress Rate (PR).
For \tangram, evaluating generative fidelity requires strictly enforcing geometric constraints. 
We introduce a hierarchical evaluation suite assessed on the final generated frame $v_T$ using a pixel-level heuristic parser (details in Appendix \ref{appsubsec:evaluator}):
\begin{enumerate}[nosep]
    \item \textit{Strict Goal Completion:} 
    A trial is successful if and only if all $N$ pieces are placed within the target silhouette without overlap, distortion, or color hallucination. This measures strict constraint satisfaction:
    $$S_{strict} = \prod_{i=1}^{N} \mathbb{I}(p_i \in g \land \text{consistent}(p_i))$$
    , where $\text{consistent}(p_i)$ ensures the piece retains its original topology (shape and color).
    \item \textit{Progress Goal Completion (Piece-wise Accuracy):} 
    To measure partial success, we compute the normalized count of correctly placed pieces within the silhouette. 
    $$S_{progress} = \frac{1}{N} \sum_{i=1}^{N} \mathbb{I}(p_i \in g \land \text{consistent}(p_i))$$

    \item \textit{Boundary Adherence (IoU):} 
    We quantify the precision of the placement by calculating the intersection over union of the generated pieces with the target silhouette area.
    $$S_{IoU} = \frac{\text{Area}(P_{gen} \cap P_{target})}{\text{Area}(P_{gen} \cup P_{target})}$$
    A score of 100 in percentage indicates perfect containment. Scores smaller than 100 indicate unfilled gaps or hallucinated pixels outside the boundary.
\end{enumerate}

\input{tables/maze_results}

\rparagraph{Models and Experiments.}
We adopt the Wan 2.2 TI2V 5B model~\citep{wan2025wan} as the primary backbone for video generation. 
Wan 2.2 TI2V 5B is a strong text-to-video diffusion model conditional on textual instruction, which is pre-trained on large-scale video data with sequences of 81 frames. 
We fine-tune only a subset of the model parameters using LoRA~\citep{hu2022lora} for 20 epochs. 
The computational cost and training details are reported in Appendix~\ref{appsec:experiment}.

In addition, we evaluate several strong proprietary and open-source MLLMs with promising visual understanding performance under zero-shot or fine-tuning settings, including GPT 5.1 \citep{openai_gpt5p1_2025}, GPT 5.2 \citep{openai_gpt5p2}, and Qwen3-VL-8B~\citep{Bai2025Qwen3VLTR}. 
For Qwen3-VL-8B, we apply full-parameter fine-tuning to support textual reasoning baselines. 
For visual reasoning, we compare against VPRL \citep{xu2025visual} on {\maze} and Qwen-Image-Edit \citep{wu2025qwenimagetechnicalreport} on \tangram, which serve as strong task-specific baselines.
All fine-tuned models are optimized using standard training objectives without task-specific auxiliary losses. 
Detailed hyperparameter configurations and prompting templates for each task and system variant are provided in Appendix~\ref{appsubsec:model_hyperparameters} and \ref{appsec:prompt}.

\subsection{In-Distribution Results}
\label{subsec:in-distribution results}
We first establish the efficacy of the proposed approach on in-distribution (IID) data, where test samples share the same distribution (maze size, icon, silhouette) as the training set.

\input{figures/tangram_result}

\rparagraph{MLLMs struggle with interpretable reasoning with interpretability.}
Proprietary MLLM consistently struggles with \maze, suggesting that while visual reasoning is trivial for humans, it remains a bottleneck for current large-scale models in line with previous work \citep{xu2025visual}.
Fine-tuning open-sourced MLLM offers limited improvement in OOD settings, with models still failing to reliably capture the underlying spatial layouts, with or without explicit coordinate supervision.
This weakness is most evident in \tangram, where text-based baselines struggle to express continuous spatial manipulations with precise coordinates, leading to overlapping pieces and near-zero Strict Goal Completion (as shown in Figure \ref{fig:tangram} with a standalone renderer).
We hypothesize this stems from the bottleneck of grounding, where MLLMs typically ground attention to semantic objects \citep{zhang2025scaling,izadi2025visual}, whereas in our context, the model grounds coordinates to empty space inside a silhouette without any semantic anchors. 
In contrast, the video generation approach explicitly simulates the reasoning process as a visual trajectory. 
By rendering the intermediate steps rather than outputting abstract coordinates, our model circumvents cross-modal grounding (visual to symbolic), offering better performance and native interpretability of failure modes, such as visual collisions, that text-based planners obscure.

\rparagraph{Video generation model enables reliable sequential planning.}
As detailed in Table \ref{tab:maze results}, the video model exhibits strong consistency in IID \maze. 
On standard $4\times4$ and $5\times5$ grids, the model attains 98.0\% and 96.0\% Exact Match (EM) scores respectively. 
These results indicate that within the training distribution, the model successfully learns to model valid spatiotemporal transitions, consistently generating trajectories that adhere to environmental constraints while successfully reaching the goal.

\input{tables/tangram_results}

\rparagraph{Visual context is essential for \tangram~as geometric priors.}
For \tangram, we introduce three system variants that vary the availability of visual context and the action types in order to examine how geometric priors affect model performance (examples in Figure \ref{appfig:tangram_tasks}):
\begin{itemize}[leftmargin=*,nosep]
    \item \textit{Fade-in}: Tangram pieces gradually appear in their target locations on an initially empty canvas, one by one.
    \item \textit{Rotation}: Pieces initially are listed outside the target silhouette with random orientations on the left side of the canvas. The model must first rotate and then translate each piece to fit the target shape, requiring both geometric transformation and spatial planning.
    \item \textit{Translation}: Pieces start outside the silhouette on the left side of the canvas, but are pre-aligned with the correct orientation. The model is required only to translate pieces into position, isolating planning from rotational reasoning.
\end{itemize}
As shown in Table~\ref{tab:tangram_results}, performance differs markedly across these variants depending on the available visual context.
In the Fade-In setting, where initial tangram shapes as geometric priors are absent, the model fails completely (0.8\% accuracy), mirroring its inability to fit the training data with $\sim0\%$ training accuracy. 
Rotation achieves intermediate performance 22.4\%, likely because while it maintains temporal context, the pixel-level warping during rotation degrades feature consistency. 
Conversely, when the visual context of the piece with more prior knowledge (shapes and orientations) is explicitly preserved via Translation, performance reaches 68.0\%. 
This gap confirms that the presence of visual context helps model to learn geometric planning with better consistency, which is further discussed in Section \ref{subsec:visual context}. 

\subsection{Zero-Shot Out-of-Distribution Generalization}
\label{subsec:ood results}
A key criticism of visual agents is their tendency to overfit to visual assets. 
We demonstrate that our model generalizes to Out-Of-Distribution (OOD) contexts without fine-tuning.

\rparagraphnospace{Generalization to OOD Maze Sizes and Path Lengths.}
We first probe the model's ability to handle unseen spatial scales. 
As detailed in Table \ref{tab:maze results}, the model generalizes robustly to spatially larger grid dimensions. 
When transitioning from seen $6\times6$ grids to unseen $7\times7$ grids, EM accuracy remains high at 92.00\%. 
While performance naturally degrades on $8\times8$ grids (78.00\%), this represents a graceful decline rather than a catastrophic failure, suggesting the model's planning horizon scales reasonably well with spatial expansion.
However, we observe a sharper performance drop when the length of the path exceeds the training distribution (temporal OOD Path Length), with accuracy falling to $36\sim42\%$. 
Crucially, rather than the reasoning capability itself, as we demonstrate in Section \ref{subsec:test time scaling}, this performance drop in either spatial or temporal OOD cases can be effectively mitigated via ``Visual Test-Time Scaling'' by increasing the generated frame count to allocate more compute for complex trajectories.

\rparagraph{Robustness for OOD Visual Agent Icon in \maze.}
To determine if the model relies on and overfits to specific visual artifacts, we replace the standard agent icon with novel, unseen patterns (OOD Icon). 
Remarkably, this visual shift induces negligible performance loss. 
On $3\times3$ mazes, accuracy shifts marginally from 94.00\% with the seen icon to 92.00\% with the OOD icon; on $5\times5$ mazes, it remains robust at 94.00\%. 
This consistency holds across all OOD maze sizes and path lengths (Table \ref{tab:maze results}, right columns).
Qualitatively (Figure \ref{appfig:maze correct showcase}), the model successfully preserves the identity and texture of the unseen icon throughout the generation, adhering to the object permanence constraint. 
These results indicate that the model has decoupled the \textit{planning algorithm} from the \textit{visual entity}; it does not simply memorize specific pixel transitions, but rather identifies the agent as a distinct entity, and applies learned movement dynamics disentangled from its visual appearance.

\rparagraph{Geometric Generalization to OOD Tangram Silhouettes.}
In the Tangram domain, we test whether the model can solve puzzles with unseen target silhouettes. 
As shown in Table \ref{tab:tangram_results}, the model exhibits strong zero-shot transfer. 
In the Translation setting, accuracy on unseen silhouettes (60.8\%) is comparable to seen silhouettes (68.0\%). 
Furthermore, even in the Rotation setting, changing the initial layout of the pieces (e.g., using the Translation test configuration) yields performance similar to the in-distribution layout.
This parity between seen and unseen geometries confirms that the model is not retrieving memorized solutions. Instead, it has internalized generalized concepts of geometric fitting and collision-free sliding, allowing it to solve novel spatial arrangement problems dynamically.

%% file: tables/maze_results.tex
\begin{table*}[t]
\centering
\caption{Results across different maze sizes and path lengths. All open-sourced models are fine-tuned and proprietary models are evaluated in a zero-shot setting. \textcolor{black!65}{\faBook} represents texts, \textcolor{black!65}{\faImage} represents images, and \textcolor{black!65}{\faFilm} represents videos. {\color{cyan!10}$\blacksquare$} represents visual reasoning systems. }
\renewcommand\arraystretch{1.3}
\resizebox{\textwidth}{!}{%
\begin{tabular}{lllcccccccccccccc}
\toprule
\multirow{4}{*}{\textbf{Model}} & \multirow{4}{*}{\textbf{Input}} & \multirow{4}{*}{\textbf{Output}} & \multicolumn{2}{c}{\textbf{In Distribution}} & \multicolumn{4}{c}{\textbf{OOD Maze Sizes}} & \multicolumn{4}{c}{\textbf{OOD Path Length}} & \multicolumn{4}{c}{\textbf{OOD Both}} \\
\cmidrule(lr){4-5} \cmidrule(lr){6-9} \cmidrule(lr){10-13} \cmidrule(lr){14-17}
& & & \multicolumn{2}{c}{3x3 -- 6x6} & \multicolumn{2}{c}{7x7} & \multicolumn{2}{c}{8x8} & \multicolumn{2}{c}{5x5 (Long)} & \multicolumn{2}{c}{6x6 (Long)} & \multicolumn{2}{c}{7x7 (Long)} & \multicolumn{2}{c}{8x8 (Long)} \\
\cmidrule(lr){4-5} \cmidrule(lr){6-7} \cmidrule(lr){8-9} \cmidrule(lr){10-11} \cmidrule(lr){12-13} \cmidrule(lr){14-15} \cmidrule(lr){16-17}
& & & \textbf{EM $\uparrow$} & \textbf{PR $\uparrow$} & \textbf{EM $\uparrow$} & \textbf{PR $\uparrow$} & \textbf{EM $\uparrow$} & \textbf{PR $\uparrow$} & \textbf{EM $\uparrow$} & \textbf{PR $\uparrow$} & \textbf{EM $\uparrow$} & \textbf{PR $\uparrow$} & \textbf{EM $\uparrow$} & \textbf{PR $\uparrow$} & \textbf{EM $\uparrow$} & \textbf{PR $\uparrow$} \\
\midrule
\noalign{\vspace{-0.7ex}}
\rowcolor{gray!10}
\multicolumn{17}{l}{\textit{Proprietary Models}} \\
GPT-5.1 & \textcolor{black!65}{\faBook + \faImage} & \textcolor{black!65}{\faBook} & 10.6 & 10.7 & 6.32 & 6.72 & 6.00 & 6.00 & 0 & 0 & 0 & 0 & 0 & 0 & 0 & 0 \\
GPT-5.2 & \textcolor{black!65}{\faBook + \faImage} & \textcolor{black!65}{\faBook} & 12.5 & 12.5 & 8.40 & 8.40 & 8.40 & 8.40 & 0 & 0 & 0 & 0 & 0 & 0 & 0 & 0 \\
\midrule
\noalign{\vspace{-0.7ex}}
\rowcolor{gray!10}
\multicolumn{17}{l}{\textit{Open-Sourced Models (All Fine-Tuned)}} \\
Qwen3-VL-8B & \textcolor{black!65}{\faBook + \faImage} & \textcolor{black!65}{\faBook} & 58.3 & 68.6 & 20.0 & 37.3 & 19.2 & 34.3 & 0 & 13.3 & 0 & 13.2 & 0 & 11.3 & 0 & 8.9 \\
\multicolumn{3}{l}{\textit{\hspace{0.4cm}- w coordinates}} & 72.0 & 77.3 & 33.2 & 45.0 & 22.0 & 30.5 & 0 & 17.1 & 0 & 13.4 & 0 & 8.1 & 0 & 5.9  \\
\rowcolor{cyan!5}VPRL-7B * & \textcolor{black!65}{\faImage} & \textcolor{black!65}{\faImage} & 73.5 & 78.6 & 14.0 & 25.2 & 4.00 & 6.20 & 0 & 11.0 & 2.00 & 16.7 & 0 & 4.10 & 0 & 0.70 \\
\rowcolor{cyan!5}Wan2.2-TI2V-5B & \textcolor{black!65}{\faBook + \faImage} & \textcolor{black!65}{\faFilm} & \textbf{96.0} & \textbf{99.0} & \textbf{90.0} & \textbf{92.3} & \textbf{80.0} & \textbf{83.6} & \textbf{44.0} & \textbf{55.2} & \textbf{42.0} & \textbf{51.6} & \textbf{40.0} & \textbf{51.1} & \textbf{32.0} & \textbf{47.1} \\
\rowcolor{cyan!5}\hspace{0.4cm}- Unseen Visual Icons &  &  & \textbf{95.5} & \textbf{98.2} & \textbf{92.0} & \textbf{92.6} & \textbf{78.0} & \textbf{81.6} & \textbf{36.0} & \textbf{46.3} & \textbf{42.0} & \textbf{52.0} & \textbf{38.0} & \textbf{47.9} & \textbf{32.0} & \textbf{42.3} \\
\bottomrule
\end{tabular}}
\label{tab:maze results}
\vspace{-2mm}
\end{table*}

%% file: figures/tangram_result.tex
\begin{figure}[t]
    \centering
    \includegraphics[width=\linewidth]{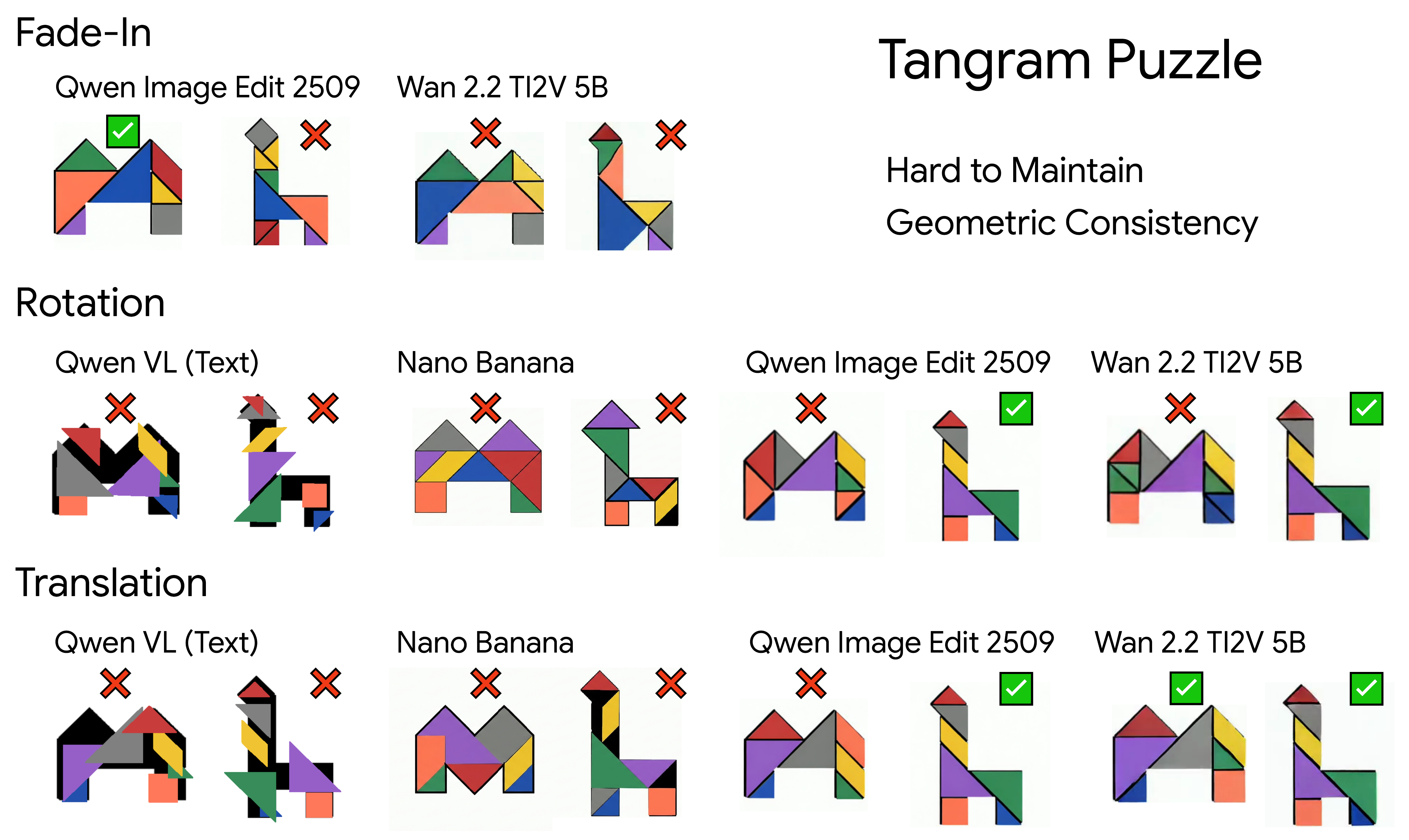}
    \caption{Generated solution of \tangram~by different system variants. For Qwen-3-VL, we visualize the layout based on the predicted coordinates and rotations. We crop the main area for the predictions from image editing model and video generation model. For video generation model, we only select the last frame as illustration here. For full details, please refer to Figure \ref{appfig:tangram showcase}. }
    \label{fig:tangram}
    \vspace{-2mm}
\end{figure}

%% file: tables/tangram_results.tex
\begin{table*}[t]
\centering
\caption{Quantitative results under Fade-In, Rotation, Translation situations for \tangram. ``GC.'' denotes goal completion, ``BA.'' denotes boundary adherence.
\textcolor{black!65}{\faBook} represents texts, \textcolor{black!65}{\faImage} represents images, and \textcolor{black!65}{\faFilm} represents videos.{\color{cyan!10}$\blacksquare$} represents visual reasoning systems. }
\renewcommand\arraystretch{1.2}
\resizebox{0.76\textwidth}{!}{%
\begin{tabular}{lcc ccc ccc}
\toprule
 & & & \multicolumn{3}{c}{Seen (\textit{Learnability})} & \multicolumn{3}{c}{Unseen (\textit{Generalizability})} \\
\cmidrule(lr){4-6} \cmidrule(lr){7-9}
Model & Input & Output & Strict GC & Progress GC & BA & Strict GC & Progress GC & BA \\
\midrule
\noalign{\vspace{-0.7ex}}
\rowcolor{gray!10} \multicolumn{9}{c}{\textit{Fade-In}} \\
\rowcolor{cyan!5}Qwen-Image-Edit-20B & \textcolor{black!65}{\faBook + \faImage} & \textcolor{black!65}{\faImage} & {31.0} & {82.3} & {99.8} & {32.0} & {81.3} & {99.7} \\
\rowcolor{cyan!5}Wan2.2-TI2V-5B & \textcolor{black!65}{\faBook + \faImage} & \textcolor{black!65}{\faFilm} & 0.80 & 49.4 & 98.1 & 0.80 & 48.9 & 98.0 \\
\midrule
\noalign{\vspace{-0.7ex}}
\rowcolor{gray!10} \multicolumn{9}{c}{\textit{Rotation}} \\
Qwen3-VL-8B & \textcolor{black!65}{\faBook + \faImage} & \textcolor{black!65}{\faBook} & 14.4 & 69.7 & 89.5 & 1.6 & 52.1 & 80.8 \\
\rowcolor{cyan!5}Nano Banana & \textcolor{black!65}{\faBook + \faImage} & \textcolor{black!65}{\faImage} & - & - & - & 9.80 & 43.4 & 64.7 \\
\rowcolor{cyan!5}Qwen-Image-Edit-20B & \textcolor{black!65}{\faBook + \faImage} & \textcolor{black!65}{\faImage} & {45.2} & {87.5} & {99.7} & {43.2} & {85.7} & {99.6} \\
\rowcolor{cyan!5}Wan2.2-TI2V-5B & \textcolor{black!65}{\faBook + \faImage} & \textcolor{black!65}{\faFilm} & 22.4 & 76.8 & 98.1 & 22.4 & 74.5 & 98.0 \\
\midrule
\noalign{\vspace{-0.7ex}}
\rowcolor{gray!10}
\multicolumn{9}{c}{\textit{Translation}} \\
Qwen3-VL-8B & \textcolor{black!65}{\faBook + \faImage} & \textcolor{black!65}{\faBook} & 28.0 & 75.7 & 91.4 & 1.60 & 58.9 & 82.4 \\
\rowcolor{cyan!5}Nano Banana & \textcolor{black!65}{\faBook + \faImage} & \textcolor{black!65}{\faImage} & - & - & - & 3.90 & 51.3 & 74.5 \\
\rowcolor{cyan!5}Qwen-Image-Edit-20B & \textcolor{black!65}{\faBook + \faImage} & \textcolor{black!65}{\faImage} & {85.7} & {97.7} & {99.9} & {76.0} & {95.4} & {99.7} \\
\rowcolor{cyan!5}Wan2.2-TI2V-5B & \textcolor{black!65}{\faBook + \faImage} & \textcolor{black!65}{\faFilm} & 68.0 & 94.7 & 97.0 & 60.8 & 92.0 & 97.0 \\
\bottomrule
\end{tabular}
}
\label{tab:tangram_results}
\vspace{-2mm}
\end{table*}

%% file: secs/5_discussion.tex
\section{Discussion and Analysis}

\input{figures/scaling_line}

\subsection{Visual Context as Control Constraints}
\label{subsec:visual context}

Rather than relying on textual instructions that are then mapped to visual patterns \citep{wan2025wan}, we posit that explicit visual context, by directly specifying visual appearance, acts as a stronger form of control for visual reasoning.

\rparagraph{Visual context outperforms textual instructions for OOD visual generalization.}
In \maze, we compared two control methods for changing the agent's appearance: 
(1) providing a visual of the new icon in the first frame at the starting point (as visual anchoring), vs. (2) providing a text description (e.g., ``a blue star''). 
The results show that the model failed to consistently generate the correct agent via text prompts, often reverting to the training distribution's default agent. 
Conversely, with visual context, the model successfully generalized to unseen icons while navigating unseen mazes.
The model effectively learns a conditional policy $\mathcal{P}(\text{Trajectory} | \text{Icon}, \text{Layout})$, treating the visual anchor as a variable to be preserved and the maze as a constraint to be navigated.
This implies that, in low-resource settings with limited training data, visual context acts as a more natural and straightforward control signal than text, enabling zero-shot adaptation to new domains.

\rparagraph{Visual context empowers geometric control for \tangram.}
Our analysis of \tangram~reveals that visual context functions as a geometric control on the generation process. 
We observe a strict performance hierarchy dictated by whether the visual context is injected, as we introduced in Section \ref{subsec:in-distribution results}. 
This hierarchy persists even when testing with baseline Image Editing models (using the first frame as input and the last frame as target), suggesting a consistent principle: visual context serves as a \textit{geometric reference for control}. 
In the Translation setting, the visual context is the strongest because the visual input provides a reference map where shape and orientation are preserved, requiring the model only to solve for location.
This suggests that high-fidelity visual context anchors the reasoning process, allowing the model to robustly apply geometric transformations within a continuous generative dynamic rather than hallucinating new shapes.

\subsection{Visual Test-Time Scaling}
\label{subsec:test time scaling}

Test-Time Compute, allowing a model to process information longer to achieve better results, has been investigated in LLMs \citep{openai_reasoning_llms_2024,muennighoff2025s1simpletesttimescaling}. 
We investigate if a parallel \textit{Visual Test-Time Scaling Law} exists for video generation for visual planning by generating more frames act as a larger inference budget for reasoning. 

\rparagraph{Scaling inference budget improves OOD generalization.}
We find that increasing the total frame count (e.g., from 81 to 101, 121 frames) improves navigation performance on both \textit{spatially} (maze size) and \textit{temporally} (path length) OOD tasks.
As shown in the first row in Figure \ref{fig:scaling line}, OOD performance increases steadily as we scale the inference budget from 61 to 121 frames.
However, we observe a ceiling effect: when scaling to 141 frames for temporal OOD cases, performance drops compared to 121 frames, though it remains superior to the training baseline of 81 frames. 
We attribute this drop to the architectural limits of the video generation model's positional embeddings \citep{wang2025deforming}, which struggle to extrapolate when the frame count deviates significantly from the training distribution.

To rigorously probe whether these gains stem from finer-grained reasoning or simply longer video duration, we introduce a control variable $\kappa$ (scaling factor), defined as the number of frames allocated per discrete step in the maze solution. 
We tested $\kappa \in \{5, 7, 9, 11\}$ across spatially and temporally OOD settings.
As shown in Figure \ref{fig:scaling line}, we observe a clear positive correlation: assigning more frames per step ($\kappa=7, 9, 11$) significantly improves performance on spatially ID and OOD settings compared to lower resolutions ($\kappa=5$). 
Notably, in temporal OOD settings, performance peaks at $\kappa=9$ before degrading at $\kappa=11$. 
This degradation aligns with the positional embedding limitation noted above, as $\kappa=11$ pushes the total video length to $\sim200$ frames. 
Crucially, this drop is \textit{not} observed in the spatially OOD setting where total path length is in-distribution, confirming that the degradation is limited by the sequence length capacity rather than logical.
These results confirm that, within architectural limits, the video generation model exhibits strong extrapolation capabilities, utilizing increased inference compute budget to resolve complex dependencies.

\rparagraphnospace{Emergent ``self-correction'' behaviors appear at higher frame budgets.~}
Beyond statistical improvements, we qualitatively observe emergent behaviors of the agent that are not included in training data when the inference budget is increased. 
In some long-horizon maze cases (see Figure \ref{appfig:maze self correction}), the model initially steers the agent toward the wrong direction. 
However, unlike in low-frame regimes, where the agent would collide with the wall, the high-frame model generates a backtracking trajectory by stopping, reversing, and correcting its path within the generated sequence. 
This suggests that the video generation is not merely retrieving a memorized path, but actively simulating the trajectory, where intermediate frames help to correct the rollout plan.

\rparagraphnospace{Fidelity-Reasoning Trade-off: Scaling is task-dependent.}
While scaling benefits sequential planning in \maze, we do not observe the same trend for \tangram. 
In the Translation setting, increasing the frame count maintains performance but yields no significant gain. 
To understand this discrepancy, we analyzed the correlation between an indicative \textit{Visual Consistency Metric} (measuring shape integrity over time, detailed introduction in Appendix \ref{appsubsec:evaluator}) and task success in Rotation and Translation, finding a Pearson correlation of $\rho\geq0.6$.
This indicates that one of the bottlenecks in \tangram~is maintaining the shape's geometry constraint during generation. 
Unlike the \maze, which moves a small icon across a static environment, tangram pieces are susceptible to geometric deformation over long generation windows with high visual change across the whole canvas. 
Therefore, while more frames help logic in \maze, they tax the model's ability to maintain high-frequency visual details in \tangram, highlighting different bottlenecks for current generative visual reasoners.

\subsection{Discussion}
\label{subsec:discussion}
\rparagraph{Generalization to Irregular Maze and Movements}
Beyond generalizing to larger grids, we observe that the video generation model exhibits remarkable zero-shot adaptation to irregular maze layouts (Figure \ref{appfig:irregular maze} in Appendix \ref{appsec:results}).
Despite being trained exclusively on grid-based environments with horizontal or vertical movement, the model successfully generates diagonal trajectories to navigate irregular, non-grid environments.
While the success rate is lower compared to standard mazes, the model still shows its potential to adhere to collision constraints without hallucinating paths through walls, even in these unseen topologies.
This indicates that the model has not simply memorized a discrete grid-based navigation, but has abstracted a policy of moving toward goal subject to collision constraints, which enables cross-domain adaptation, and we call for further research.

\rparagraph{Video Generation v.s. Image Editing}
We observe that while Image Editing achieves higher performance on the \tangram~task, Video Generation offers better interpretability and verifiability by producing the intermediate reasoning traces.
We attribute the performance gap in Tangrams to two primary factors. 
In addition to the model sizes, where the image editing model is twice as large as the video generation model, another potential factor is optimization density. 
Video generation model requires modeling the joint probability of the entire sequence $P(v_0,...,v_T)$, which requires more effort to fit every frames during training.  
In contrast, the image editing model optimizes only the conditional marginal $P(s_{goal}|s_{start})$, allowing for full capacity for optimizing final visual fidelity without being distracted by temporal consistency maintenance. 

%% file: figures/scaling_line.tex
\begin{figure*}[htbp]
    \centering
    \includegraphics[width=\linewidth]{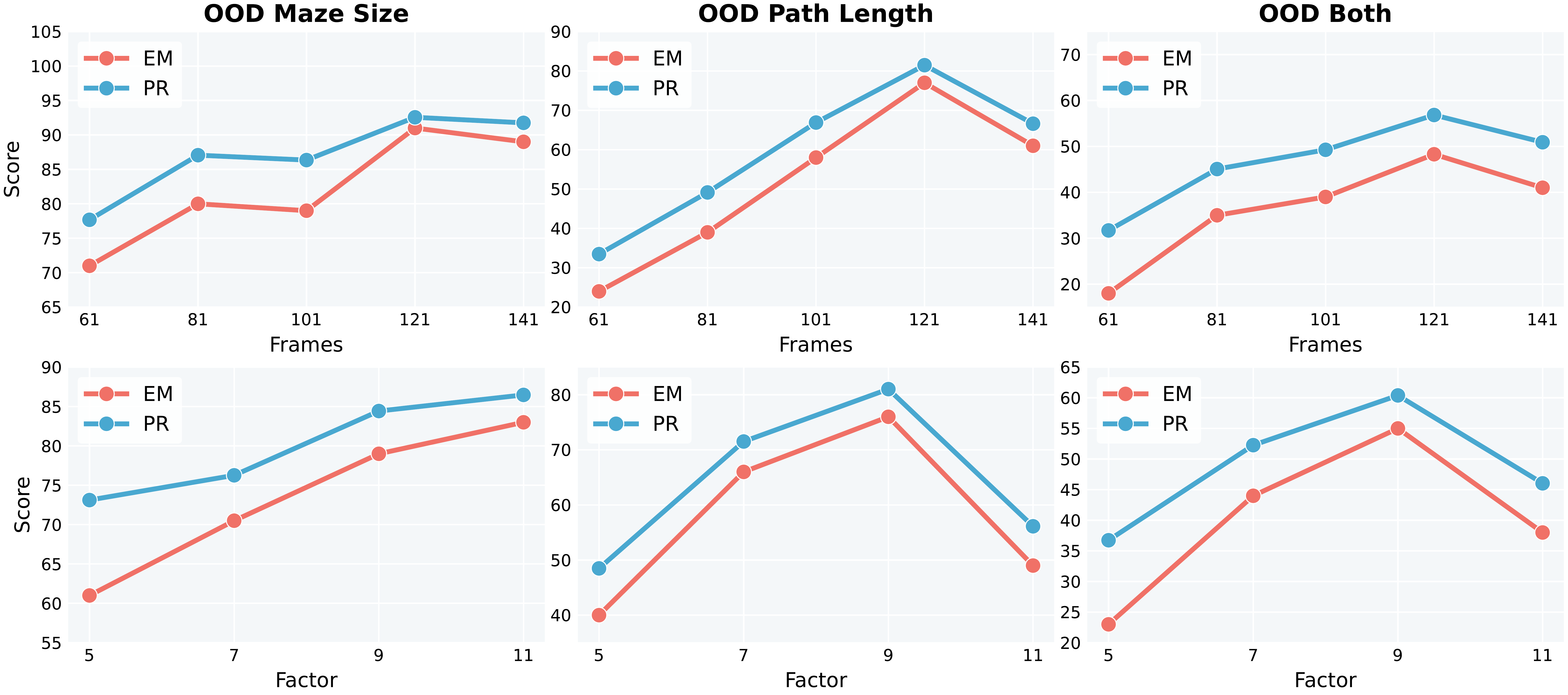}
    \caption{Visual Test-Time Scaling for \maze~using unseen icon with more inference budget. Row 1 shows the performance curve when increasing the total number of frames per video; Row 2 shows the performance curve when changing the scaling factor $\kappa$ to allocate a different number of frames per discrete step in the maze solution. Detailed results for both settings are shown in Figure \ref{appfig:maze scaling frames} and \ref{appfig:maze scaling kappa}. }
    \label{fig:scaling line}
\end{figure*}

%% file: secs/6_conclusion.tex
\section{Conclusion}
In this work, we demonstrate that video generation models are not merely for visual synthesis but also as powerful engines for visual reasoning. 
Through systematic investigations on the discrete logical challenges of {\maze} and the continuous, high-fidelity geometric constraints of {\tangram}, we demonstrate that generative video affords an expressive representation for spatial planning that text-based MLLMs inherently lack. 
Our results show that video-based planners exhibit strong generalization, effectively disentangling task-level logic from superficial visual patterns. 
Beyond other performance bottlenecks, such as maintaining geometric consistency, we further identify a visual test-time scaling phenomenon: as the number of generated frames, serving as a temporal reasoning budget, increases, models display emergent competence on out-of-distribution, long-horizon sequential problems.
This behavior mirrors the transition to powerful System-2-style reasoning observed in Large Language Models, which we call for further research.

%% file: secs/impact_statement.tex

%% file: appendix/A_experiment_setups.tex
\section{Experimental Setups}
\label{appsec:experiment}

\subsection{Dataset}
\label{appsubsec:dataset}

\paragraph{\maze}
We evaluate models across a suite of grid-world maze navigation environments \citep{ivanitskiy2023configurable}, with dimensions ranging from $3\times3$ to $8\times8$ and optimal planning lengths spanning $2$--$18$ steps. 
Each configuration features a unique spatial layout, with the agent's initial state $v_0$ sampled from reachable grid locations relative to the goal.
We follow the practice in \citet{xu2025visual} where the agent location is randomly initialized, and the solution is obtained from the built-in searching algorithm from \citet{ivanitskiy2023configurable}. 

For the training set, we generate $1{,}000$ unique instances for maze sizes from $3 \times 3$ to $6 \times 6$, and maximum navigation path length $\leq 12$.
To ensure visual diversity and complexity, we use $40$ unique visual icons instead of the unique one icons, as shown in Figure \ref{appfig:maze icon}. 
It is worth noting that we do not explicitly replicate maze layouts with different visual icons; consequently, the same environment does not appear with multiple agent icon variants.
To assess the model's robustness and systematic generalization, we compile the held-out evaluation set into three difficulty tiers based on structural and temporal complexity (see Table~\ref{apptab:maze_stats}):
\begin{itemize}
    \item \textit{In-Distribution:} Test cases share the same distribution of grid sizes and planning lengths as the training set, featuring novel maze layouts and randomized start/goal configurations.
    \item \textit{OOD Maze Sizes (Spatial OOD):} Environments scale to $7\times7$ and $8\times8$ dimensions while maintaining distribution of training-level navigation path steps, evaluating adaptability to unseen spatial scales.
    \item \textit{OOD Path Length (Temporal OOD):} We increase the path length as the indication of reasoning complexity to $13$--$18$ steps across in-distribution grid sizes ($5\times5$, $6\times6$), specifically challenging long-range dependency modeling and compositional reasoning.
    \item \textit{OOD Both (Temporal \& Spatial OOD):} We increase both maze sizes and path length to $13$--$18$ steps across grid sizes from $7\times7$ to $8\times8$. 
\end{itemize}

In parallel, we construct a visually OOD test set featuring icons that are unseen during training (right panel of Figure \ref{appfig:maze icon}). 
The instance-wise maze layouts are identical across icon sets, enabling a controlled evaluation of the model’s ability to generalize to unseen visual appearances.

\input{figures/appfig_maze_icon}

\begin{table}[ht]
\centering
\caption{Statistics of {\maze} across types}
\label{apptab:maze_stats}
\begin{tabular}{llccc}
\toprule
\textbf{Type} & \textbf{Category} & \textbf{Grid Sizes} & \textbf{Path Length} & \textbf{Numbers} \\ \midrule
\textbf{Train set} & - & $3\times3$ to $6\times6$ & $2$ -- $12$ steps & 4,000 $(4 \times 1,000)$ \\ \midrule
\textbf{Test set} & In-Distribution & $3\times3$ to $6\times6$ & $2$ -- $12$ steps & 1,000 $(4 \times 250)$  \\
 & Spatial OOD & $7\times7$ to $8\times8$ & $2$ -- $12$ steps & 500 $(2 \times 250)$  \\
 & Temporal OOD & $5\times5$ to $6\times6$ & $13$ -- $18$ steps & 500 $(2 \times 250)$  \\ 
  & Spatiotemporal OOD & $7\times7$ to $8\times8$ & $13$ -- $18$ steps & 500 $(2 \times 250)$  \\
 \bottomrule
\end{tabular}
\end{table}

\paragraph{\tangram}
We introduce a Tangram-based puzzle-solving task derived from the Kilogram dataset \citep{ji-etal-2022-abstract}. To construct our dataset, we sample all black-background silhouettes from the training set of Kilogram and remove the underlying grid structures, for a total of 692 data items. For each silhouette we provide seven disjoint, color-coded geometric pieces, and every silhouette has a unique ground-truth layout. 
Based on different degrees of geometric transformation and whether to provide visual context, we introduce three variants of \tangram: Fade-In, Rotation, and Translation, as shown by Figure \ref{appfig:tangram_tasks}. 
\begin{itemize}
    \item \textit{Fade-In:} Pieces are initialized in their canonical orientations. The model's objective is limited to localizing target centroids within the silhouette.
    \item \textit{Rotation:} Pieces are initialized with random orientations, listed on the left sidebar. The model must perform sequential $\mathrm{SO}(2)$ rotations followed by spatial translations to reconstruct the target silhouette.
    \item \textit{Translation:} Pieces are initialized in their canonical orientations in the left sidebar. The model's objective is limited to localizing target centroids within the silhouette.
\end{itemize}

For evaluation, we use the held-out split of the Kilogram dataset (125 data items) to ensure zero layout overlap of silhouette patterns between training and test distributions.


\input{figures/appfig_tangram_task}

\subsection{Models and Hyper-Parameters}
\label{appsubsec:model_hyperparameters}
To investigate the impact of modalities on planning performance, we perform supervised fine-tuning (SFT) across three distinct modalities: text, image, and video.
\paragraph{Planning in text.}
In this formulation, the model leverages natural language to represent discrete action sequences in \maze~and use coordinates and rotation degrees in JSON format for \tangram. 
Formally, given a visual state $v \in \mathcal{V}$ and a task-oriented textual prompt $p$, the model is trained to generate a textual plan $t=(t_1,\dots,t_L)$, where each token $t_i$ belongs to the language vocabulary $\mathcal{V}_{\mathrm{text}}$.
The input is constructed by concatenating the visual tokens with the prompt tokens. 
Following standard protocols for autoregressive supervised fine-tuning \citep{wei2022finetuned}, we optimize model parameters $\theta$ by minimizing the negative log-likelihood of the target action sequence:
\begin{equation}
\mathcal{L}_{\mathrm{SFT}}(\theta)
= -\mathbb{E}_{(v,p,t)\sim\mathcal{D}}\left[\sum_{i=1}^{L}\log \pi_{\theta}\big(t_i \mid t_{<i}, v, p\big)\right]
\end{equation}
where $\mathcal{D}$ denotes the training distribution. This formulation treats planning as a conditional sequence-generation task, grounding the language-based action space in the visual context.

For implementation, we evaluate proprietary models GPT-5.1 \citep{openai_gpt5p1_2025} and GPT-5.2 \citep{openai_gpt5p2}\footnote{We utilize GPT-5.1 (2025-11-13) and GPT-5.2 (2025-12-11) via the Azure API.} on \maze.
For open-source baselines, we fine-tune Qwen 2.5 VL 7B \citep{bai2025qwen2} with full parameter updates on both tasks.
For \tangram, the Qwen model is trained to output 2D center coordinates and floating-point rotation degrees in a strict JSON format (prompt templates in Appendix \ref{appsec:prompt}).

\paragraph{Planning in image.}

For \maze, we adopt the Visual Planning framework \citep{xu2025visual}, where the solution is generated as a sequence of discrete images (one frame per step). 
To ensure a rigorous baseline, we strictly replicate the architecture, reward design, and single-icon (blue star) training distribution from \citet{xu2025visual}. We restrict this comparison to \maze~to avoid implementation biases that might arise from adapting their specialized reward model to \tangram.

For \tangram, we treat planning as a ``direct prediction'' problem using Image Editing. 
We first evaluate the proprietary model (Nano-Banana, formally, Gemini 3 Pro Image \citep{deepmind_gemini3proimage_2025}).
Due to the financial constraint, we only evaluate the first 51 instances in \tangram~test set with Nano Banana. 
For open-sourced models, we also experiment with Qwen Image Edit 2509 \citep{wu2025qwenimagetechnicalreport} (20B parameters), a state-of-the-art open-source image editing model. 
We fine-tune the model using LoRA \citep{hu2022lora} with default hyperparameters. Input resolutions are set to $256\times256$ for the Fade-In setting and $256\times512$ for Rotation/Translation settings to accommodate the sidebar workspace.

\paragraph{Planning in video.}
We use Wan 2.2 TI2V 5B as the main model backbone for experiments, which is a text-to-video diffusion model conditional on textual instructions. 
It's pretrained on large-scale video data, each with sequences of 81 frames. 
For experiments, we fine-tune the model with LoRA \citep{hu2022lora} for 20 epochs for both \maze~and \tangram. 
Specifically, in \maze, we train the model on 81-frame sequences, aligned with the pre-trained video length, with a fixed resolution of $480\times832$. 
In \tangram, preliminary experiments revealed that compressing the complex rotation and translation of Tangram pieces into 81 frames resulted in rapid motion blurring, which degraded geometric consistency. 
Apart from adhering to the pre-trained video length in the Fade-In setting, to mitigate this, we extended the sequence length to 201 frames for the Rotation setting.
For Translation, we employed random frame lengths (sampled from 61 to 81 with a stride of 4) as data augmentation to improve robustness.

All diffusion models are trained using the DiffSynth-Studio framework\footnote{\url{https://github.com/modelscope/DiffSynth-Studio}}. 
Detailed hyperparameters are provided in Table \ref{apptab:hyperparameters}.


\begin{table}[t]
\centering
\caption{Hyperparameters for supervised fine-tuning across different modalities.}
\label{apptab:hyperparameters}
\renewcommand{\arraystretch}{1.15}
\begin{tabular}{lccc}
\toprule
\textbf{Hyperparameter} & \textbf{Text} & \textbf{Image} & \textbf{Video} \\
\midrule
Epochs                 & 20            & 20             & 20             \\
Learning rate          & $1\times10^{-5}$ & $1\times10^{-4}$ & $1\times10^{-4}$ \\
Train batch size       & 16 & 1 & 1 \\
Gradient accumulation  & 1             & 1              & 1              \\
GPUs                   & 2             & 1              & 1              \\
\midrule
LoRA rank              & 128            & 32             & 32             \\
LoRA modules           & 
\begin{tabular}[t]{@{}l@{}}
\texttt{q\_proj,k\_proj,v\_proj} \\
\texttt{o\_proj, gate\_proj,} \\
\texttt{down\_proj, up\_proj}
\end{tabular}
&
\begin{tabular}[t]{@{}l@{}}
\texttt{to\_q, to\_k, to\_v, to\_out.0} \\
\texttt{add\_q\_proj, add\_k\_proj} \\
\texttt{add\_v\_proj, to\_add\_out} \\
\texttt{img\_mlp.net.2, img\_mod.1} \\
\texttt{txt\_mlp.net.2, txt\_mod.1}
\end{tabular}
&
\begin{tabular}[t]{@{}l@{}}
\texttt{q, k, v, o} \\
\texttt{ffn.0, ffn.2}
\end{tabular}
\\
\bottomrule
\end{tabular}
\end{table}


\subsection{Evaluator}
\label{appsubsec:evaluator}
\paragraph{\maze}

To evaluate the reasoning accuracy across diverse agent appearances and stochastic generation speeds, we designed a pipeline prioritizing visual robustness and temporal invariance based on the implementation from \citet{miniveo3reasoner}. First, to handle varying agent designs (different icons) without training specific detectors, we adopt a motion-centric extraction approach. By modeling the static maze environment as a reference, we isolate the agent solely based on its dynamics via background subtraction. To further mitigate generative artifacts, such as flickering or hallucinated objects, we impose spatiotemporal continuity constraints, restricting the tracker to physically plausible local neighborhoods. Second, addressing the variable pacing inherent in diffusion models, we reject frame-level comparisons ($p_t$ vs. $p^{gt}_t$) as they are sensitive to speed mismatches. Instead, we implement a speed-invariant alignment protocol. By resampling trajectories based on cumulative geometric distance rather than time indices, we decouple spatial path fidelity from temporal dynamics. This ensures the metric rigorously evaluates where the agent went, regardless of how fast it travelled relative to the ground truth. Based on the implementation above, we adopt the evaluation metric (EM for Exact Match and PR for Progress Rate) from \citet{xu2025visual}. 

\paragraph{\tangram}
To rigorously assess the generative model's reasoning capabilities without bias, we developed a deterministic, rule-based visual evaluator. 
Unlike neural-based evaluators (e.g., CLIPScore \citep{hessel2021clipscore}), which correlate poorly with precise geometric constraints, our pipeline operates on pixel-level segmentation and geometric primitives.
The evaluation pipeline processes the generated video $V = \{v_0, \dots, v_T\}$ and extracts metrics based on color segmentation, contour approximation, and temporal consistency.


We define the set of $N=7$ tangram pieces based on a predefined palette of unique colors $C = \{c_1, \dots, c_N\}$, where for Fade-In, the colors are fixed in the whole dataset, for Translation and Rotation, the colors are randomly assigned to each tangram pieces as they are listed on the left side of the image as visual context.

We first establish ground-truth physical geometric properties for each piece $i$. 
For Fade-In, the geometric properties of the tangram pieces could be obtained from the golden target layouts; for Translation and Rotation, the geometric properties are obtained from the left side of the first frame (input frame). 
The properties include: 
\begin{itemize}[leftmargin=*,nosep]
    \item Initial Area ($A_i^{(0)}$): The pixel count of the color mask in the source region.
    \item Initial Shape Class ($S_i^{(0)}$): The geometric topology (triangle, square, parallelogram) identified via contour approximation.
    \item Target Silhouette ($M_{tgt}$): The binary mask of the black target shape in the target region.
\end{itemize}
Specifically, to identify shapes under potential generation artifacts (e.g., anti-aliasing or slight warping), we use \texttt{cv2.approxPolyDP} to approximate the contour of each segmented piece into vertices.
We classify the shape $S_i$ based on the number of vertices, internal angles, and aspect ratios:
\begin{itemize}[leftmargin=*,nosep]
    \item Triangle: 3 vertices.
    \item Square: 4 vertices, aspect ratio $\approx 1.0$, and all internal angles $\theta \in [85^\circ, 95^\circ]$.
    \item Parallelogram: 4 vertices, opposite sides equal in length, and opposite internal angles equal (with a tolerance of $\pm 15^\circ$).
\end{itemize}

Based on the implementation above, we define the metrics in Section \ref{subsec:regimes} as follows. 

\textit{Goal Completion}~~
This metric evaluates the final state of the puzzle in the last frame $v_T$. 
A trial is considered successful if the average piece-wise completion score is 1.0.
For each piece $i$, the completion score $u_i \in [0, 1]$ is calculated based on following constraints:
(1) \textit{Area Consistency}: The piece must be present in the target region with an area $A_i^{(T)}$ within a tolerance window of the initial area to account for rotation-induced aliasing:
$0.6 \le \frac{A_i^{(T)}}{A_i^{(0)}} \le 1.4$.
(2) \textit{Shape Preservation}:
The classified shape class must match the ground truth: $S_i^{(T)} = S_i^{(0)}$.
(3) \textit{Within the silhouette}: The shape must be within the target silhouette. 
Only if the piece satisfies the three constraints above would it be considered correct. 

\textit{Boundary Adherence (Mask IoU)}~~
This metric quantifies how well the generated pieces fit inside the target silhouette without hallucinating pixels outside the lines.
We compute the union of all generated color masks in the final frame, $M_{gen} = \bigcup_{i=1}^N M_i^{(T)}$.
The score is defined as the pixel-wise precision relative to the target silhouette mask $M_{tgt}$:
$$S_{IoU} = \frac{|M_{gen} \cap M_{tgt}|}{|M_{gen} \cup M_{tgt}|}$$
A score of 1.0 implies the generated puzzle perfectly fills the silhouette with zero overflow and zero gaps.

\textit{Indicative Visual Consistency (Piece Integrity)}~~
To measure whether the model respects ``object permanence'' throughout the video (rather than pieces vanishing or flickering), we compute a temporal integrity score.
We sample frames evenly across the video sequence. For each sampled frame $v_t$, we calculate the ratio of total visible area for each color (summed across both source and target regions) relative to the initial area.
Let $r_{i,t} = A_i^{(t)} / A_i^{(0)}$. 
We define a binary integrity flag $\delta_{i,t} = \mathbb{I}(0.6 \le r_{i,t} \le 1.4)$.
The final integrity score is calculated by comparing the sequence of integrity flags against a reference ground-truth video (if available) or the ideal static assumption, quantifying the fraction of frames where all pieces maintain physical consistency.
To be noted that this metric only serves as an indicative metric for analysis. Therefore, it is not included in the main result table.

%% file: figures/appfig_maze_icon.tex
\begin{figure*}[h]
    \centering
    \includegraphics[width=\linewidth]{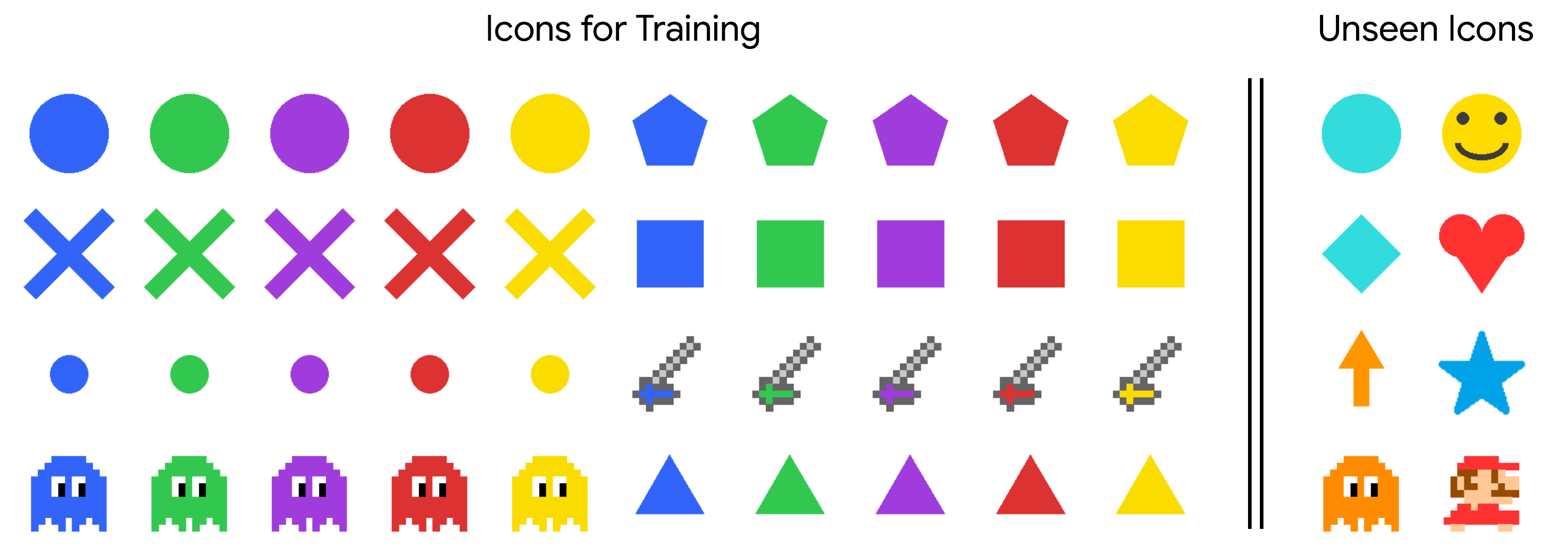}
    \caption{Agent Icons for \maze~during training and visual OOD evaluation.}
    \label{appfig:maze icon}
\end{figure*}

%% file: figures/appfig_tangram_task.tex
\begin{figure*}[t]
    \centering
    \includegraphics[width=\linewidth]{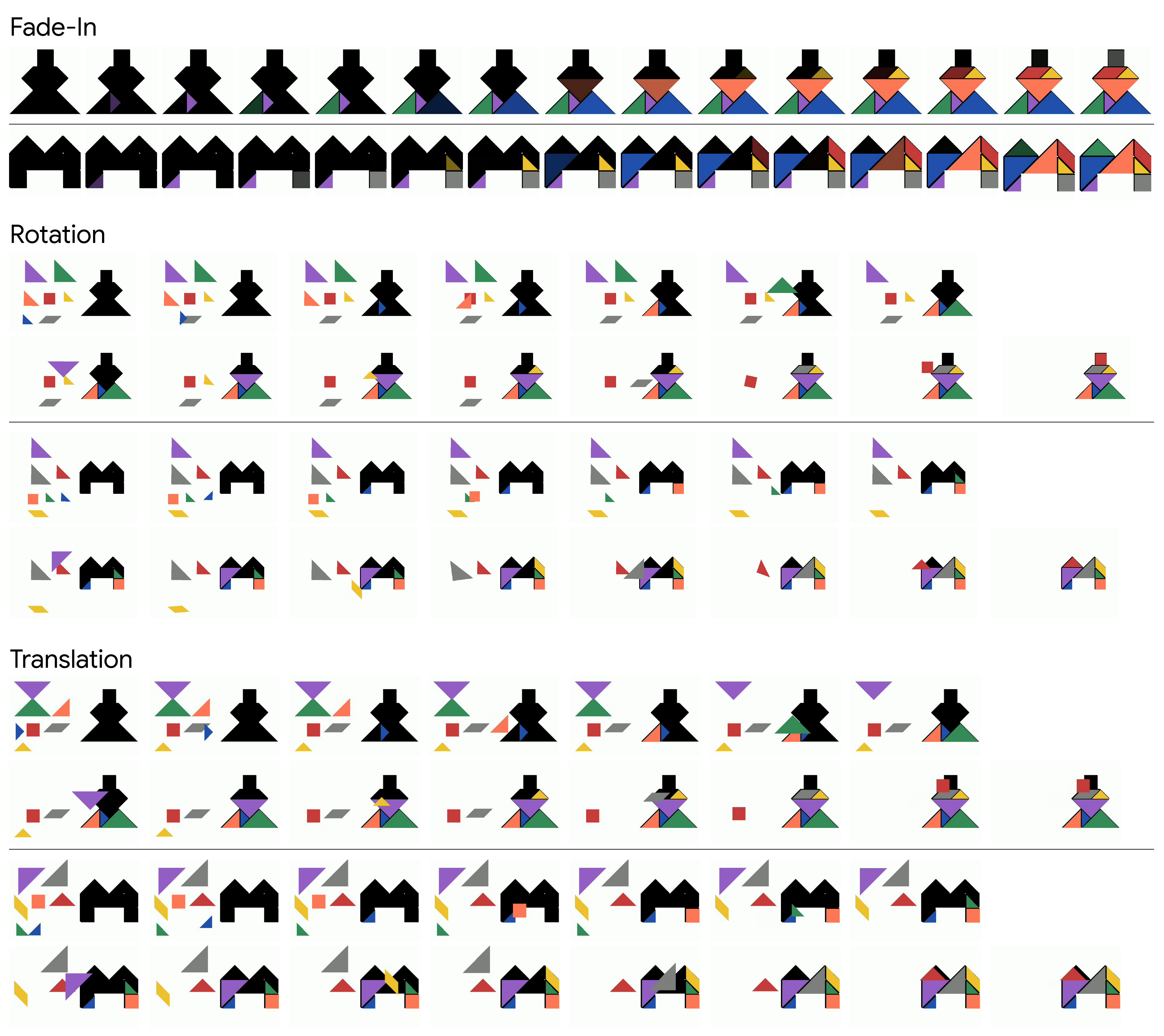}
    \caption{Illustration of different variants for \tangram. }
    \label{appfig:tangram_tasks}
\end{figure*}

%% file: appendix/B_more_results.tex
\section{Results}
\label{appsec:results}

\paragraph{Fine-Grained Results for \maze.}
Supplement to Table \ref{tab:maze results}, we illustrate the fine-grained in-distribution performance gap between general proprietary models and specialized finetuned MLLMs in \maze~tasks in Table~\ref{tab:maze_seen_results}. 
A consistent inverse correlation is observed between maze complexity and model performance; as grid size scale from 3×3 to 6×6, all models exhibit a systematic degradation in EM and PR scores, highlighting the inherent difficulty of larger maze planning. 
Notably, the proprietary GPT-5 series struggles to surpass a $20\%$ success rate. 
In contrast, Wan2.2 TI2V 5B model achieves near-saturated performance (up to $99.98\%$ PR), demonstrating that internalizing spatial dynamics through video-based output modalities provides a superior inductive bias for such visual-first navigation task. 

\input{tables/maze_seen_results}

\paragraph{Qualitative Analysis for \maze.}
Figure \ref{appfig:maze correct showcase} and Figure \ref{appfig:maze wrong showcase} show the correct and wrong generations from video generation model in \maze. 
We observe that Wan 2.2 TI2V 5B demonstrates robust object permanence throughout the generated video with high accuracy for both seen icons and unseen icons across ID and OOD settings. 
This underscores a robust zero-shot generalization capability that transcends simple pixel-level memorization, positioning multimodal generative priors as a more viable backbone for complex visual-first reasoning tasks than traditional MLLMs. 
To quantitatively analyze the failure cases, we categorize observed failures into distinct taxonomies. 
For {\maze}, we identify four primary failure modes: 1) boundary violation (crossing the walls), 2) structural distortion (unstable generation of visual icons), 3) kinematic inconsistency (agent disappearance or teleportation), and 4) wrong movement actions.
Qualitatively, we observe that similar visual semantics (e.g., similar colors) between agent icon and destination icon are more prone to error such as kinematic inconsistency, or even a minor case where the model confuses the agent icon with the goal icon, indicating the importance of visual semantics.

\paragraph{Qualitative Analysis for \tangram.}
Figure \ref{appfig:tangram showcase} shows the visualized predictions from different systems in different settings. 
For Qwen VL with textual reasoning, the model fails to output precise continuous coordinates, yielding zero strict goal completion, although being specifically tuned. 
This indicates that the text modality struggles to describe continuous fine-grained rotation and translation, although it doesn't need to maintain the geometric consistency to manipulate the pieces with coordinates. 
For image editing model, as discussed in Section \ref{subsec:discussion}, we show that after being supervised fine-tuned on the corresponding tasks, it can achieve much better completion results in \tangram~compared to proprietary models which struggles to maintain the shape of tangram pieces, as shown in Figure~\ref{appfig:tangram showcase}. 
Fine-tuned video generation model excels in the Translation setting but struggles to maintain the geometric properties of tangram pieces in Fade-In and Rotation.
We identify the primary failure modes by geometric and visual fidelity as follows: chromatic distortion (pieces change color and shape during the process, which is most evident for visual generative models), centroid displacement (the piece is placed away from the target location, which is most evident for textual reasoning models), and angular deviation (incorrect rotational orientation). 
These failure cases are qualitatively illustrated in Figures~\ref{appfig:tangram showcase}. 

\input{figures/appfig_unseen_icon_frames}
\input{figures/appfig_unseen_icon_factors}

\input{figures/appfig_maze_self_correction}

\input{figures/appfig_irregular_maze}

\input{figures/appfig_tangram_scaling}

Our error analysis reveals distinct bottlenecks across both domains. Although video generation models can adapt to physical environment rules (e.g., not crossing walls) in \maze, they struggle to maintain structural fidelity of visual semantics. This issue manifests in both \maze, where changes in visual semantics often lead to errors, and \tangram, where geometric distortions accumulate over time in the video. These findings suggest that, despite seemingly strong agent-level reasoning, sustaining long-term consistency of visual semantics remains a fundamental challenge for visual generative models when used as visual reasoners. This limitation has been largely obscured by prior work that focuses primarily on local visual changes, and we therefore call for further research in this direction.

\input{figures/appfig_maze_correct_showcase}

\input{figures/appfig_maze_error_showcase}

\input{figures/appfig_tangram_showcase}

%% file: tables/maze_seen_results.tex
\begin{table*}[t]
\centering
\caption{Evaluation results of various models across different maze sizes and path lengths. \textcolor{black!65}{\faBook} represents texts, \textcolor{black!65}{\faImage} represents images, and \textcolor{black!65}{\faFilm} represents videos. {\color{cyan!10}$\blacksquare$} represents visual reasoning systems. }
\renewcommand\arraystretch{1.2}
\resizebox{0.8\textwidth}{!}{%
\begin{tabular}{lllrrrrrrrr}
\toprule
\multirow{4}{*}{\textbf{Model}} & \multirow{4}{*}{\textbf{Input}} & \multirow{4}{*}{\textbf{Output}} & \multicolumn{2}{c}{3x3} & \multicolumn{2}{c}{4x4} & \multicolumn{2}{c}{5x5} & \multicolumn{2}{c}{6x6} \\
\cmidrule(lr){4-5} \cmidrule(lr){6-7} \cmidrule(lr){8-9} \cmidrule(lr){10-11}
& & & \textbf{EM $\uparrow$} & \textbf{PR $\uparrow$} & \textbf{EM $\uparrow$} & \textbf{PR $\uparrow$} & \textbf{EM $\uparrow$} & \textbf{PR $\uparrow$} & \textbf{EM $\uparrow$} & \textbf{PR $\uparrow$} \\
\midrule
\noalign{\vspace{-0.7ex}}
\rowcolor{gray!10}
\multicolumn{11}{l}{\textit{Proprietary Models}} \\
GPT-5.1 & \textcolor{black!65}{\faBook + \faImage} & \textcolor{black!65}{\faBook} & 15.6 & 16.0 & 11.6 & 11.6 & 8.40 & 8.40 & 6.80 & 6.80  \\
GPT-5.2 & \textcolor{black!65}{\faBook + \faImage} & \textcolor{black!65}{\faBook} & 18.4 & 18.4 & 13.2 & 13.2 & 10.0 & 10.0 & 8.40 & 8.40  \\
\midrule
\noalign{\vspace{-0.7ex}}
\rowcolor{gray!10}
\multicolumn{11}{l}{\textit{Open-Sourced Models}} \\
Qwen2.5-VL-7B & \textcolor{black!65}{\faBook + \faImage} & \textcolor{black!65}{\faBook} & 88.8 & 91.5 & 62.0 & 69.6 & 45.2 & 57.6 & 23.6 & 39.0  \\
\multicolumn{3}{l}{\textit{\hspace{0.4cm}- w CoT}} & 83.6 & 85.8 & 53.6 & 61.8 & 40.8 & 51.9 & 24.0 & 36.4  \\
Qwen3-VL-8B & \textcolor{black!65}{\faBook + \faImage} & \textcolor{black!65}{\faBook} & 89.2 & 91.5 & 69.6 & 76.7 & 44.8 & 60.0 & 29.6 & 46.2  \\
\multicolumn{3}{l}{\textit{\hspace{0.4cm}- w CoT}} & 94.4 & 94.6 & 79.6 & 83.5 & 64.0 & 70.8 & 50.0 & 60.4  \\
\rowcolor{cyan!5}VPRL-7B * & \textcolor{black!65}{\faImage} & \textcolor{black!65}{\faImage} & 94.0 & 96.0 & 72.0 & 76.0 & 66.0 & 74.4 & 62.0 & 68.0 \\
\rowcolor{cyan!5}Wan2.2-TI2V-5B & \textcolor{black!65}{\faBook + \faImage} & \textcolor{black!65}{\faFilm} & \textbf{96.0} & \textbf{99.8} & \textbf{98.0} & \textbf{99.4} & \textbf{98.0} & \textbf{99.9} & \textbf{92.0} & \textbf{94.8}  \\
\rowcolor{cyan!5}\hspace{0.4cm}\textit{- Unseen Visual Icons} &  &  & \textbf{94.0} & \textbf{98.6} & \textbf{98.0} & \textbf{99.4} & \textbf{96.0} & \textbf{99.9} & \textbf{94.0} & \textbf{94.8} \\
\bottomrule
\end{tabular}}
\label{tab:maze_seen_results}
\end{table*}

%% file: figures/appfig_unseen_icon_frames.tex
\begin{figure*}[t]
    \centering
    \includegraphics[width=\linewidth]{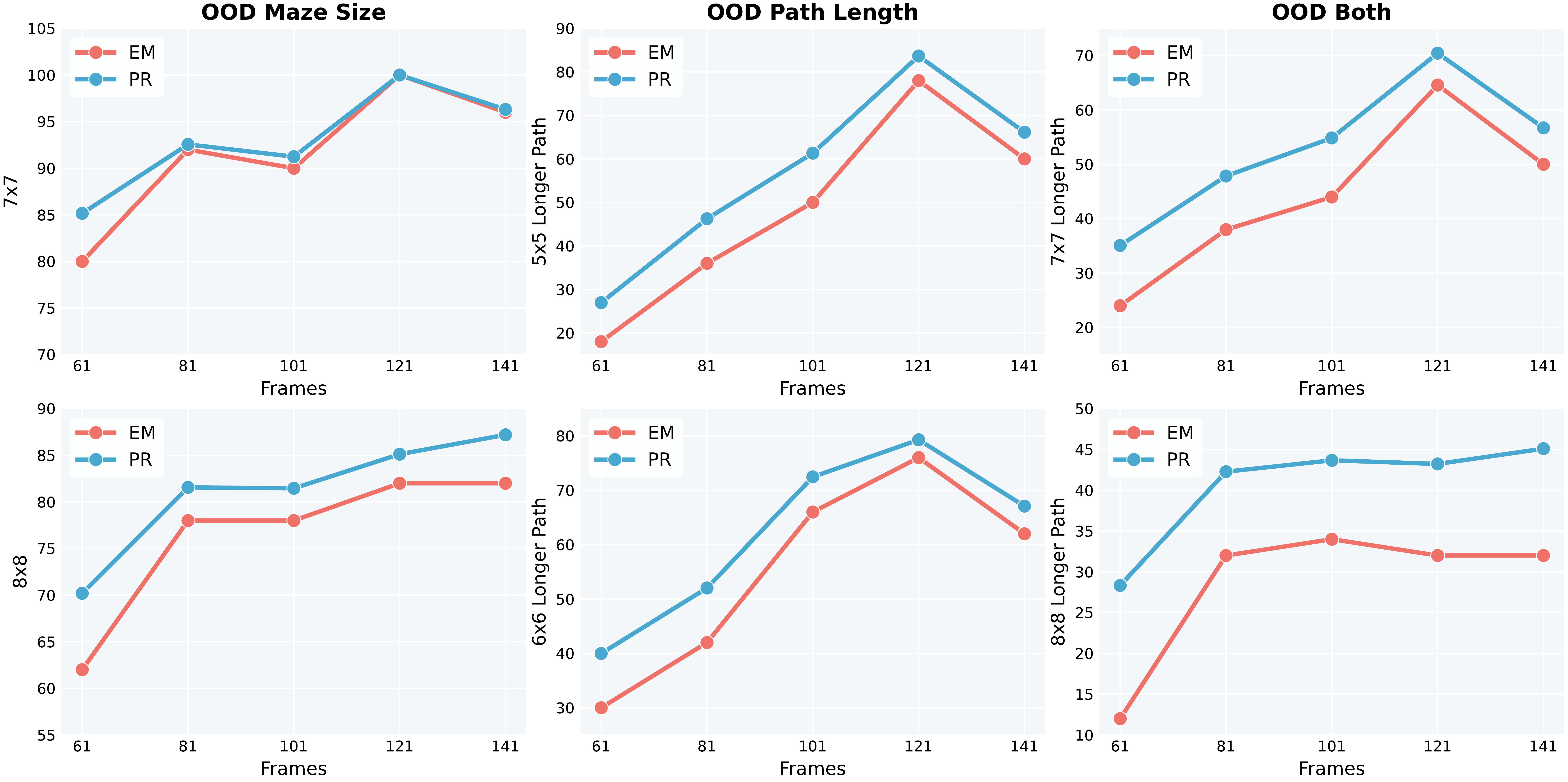}
    \caption{Fine-Grained Visual Test-Time Scaling for \maze~using unseen icon with more inference frame budget.}
    \label{appfig:maze scaling frames}
\end{figure*}

%% file: figures/appfig_unseen_icon_factors.tex
\begin{figure*}[t]
    \centering
    \includegraphics[width=\linewidth]{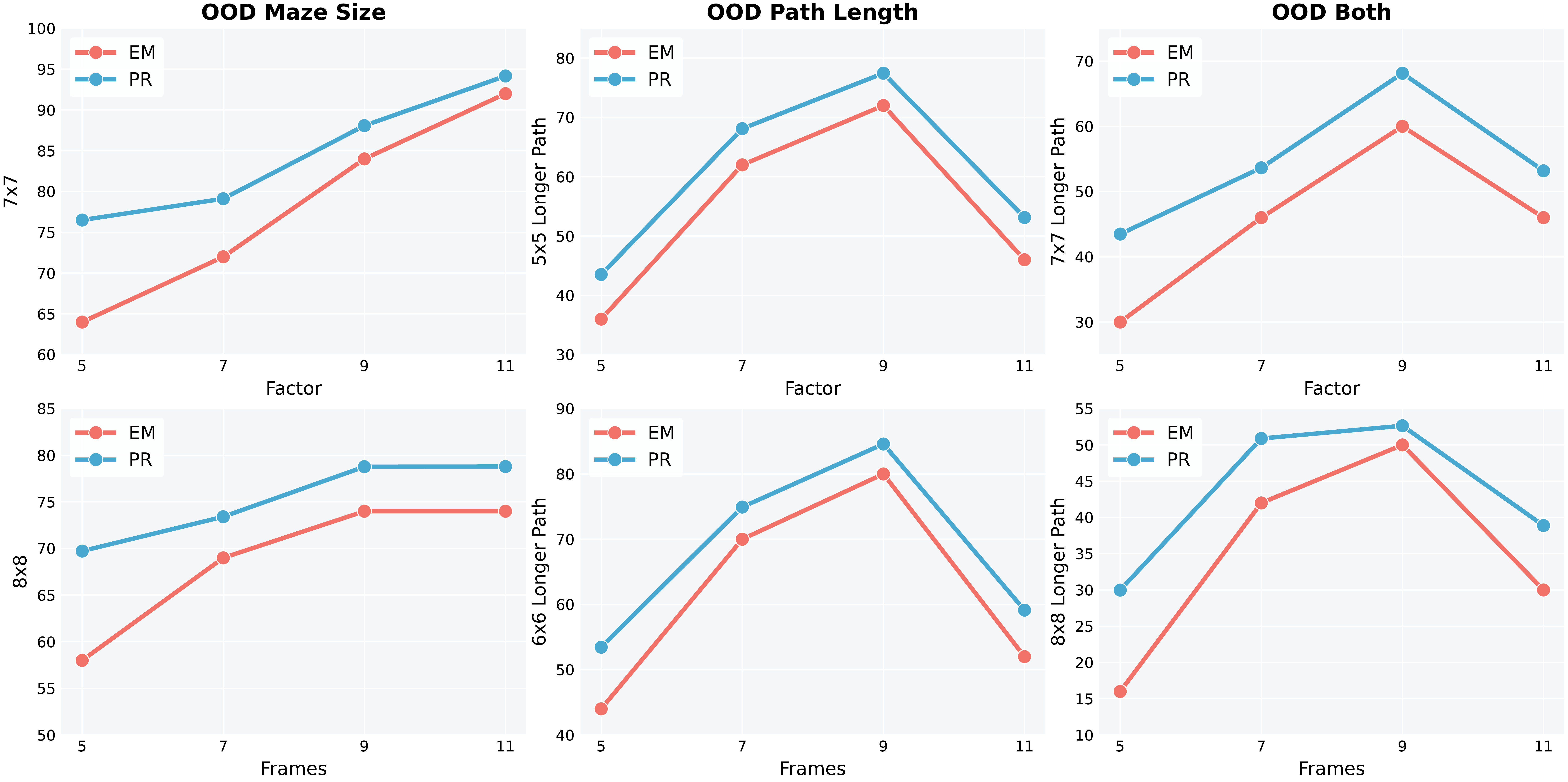}
    \caption{Fine-Grained Visual Test-Time Scaling for \maze~using unseen icon with more inference budget determined by the control variable scaling factor $\kappa$.}
    \label{appfig:maze scaling kappa}
\end{figure*}

%% file: figures/appfig_maze_self_correction.tex
\begin{figure*}[t]
    \centering
    \includegraphics[width=\linewidth]{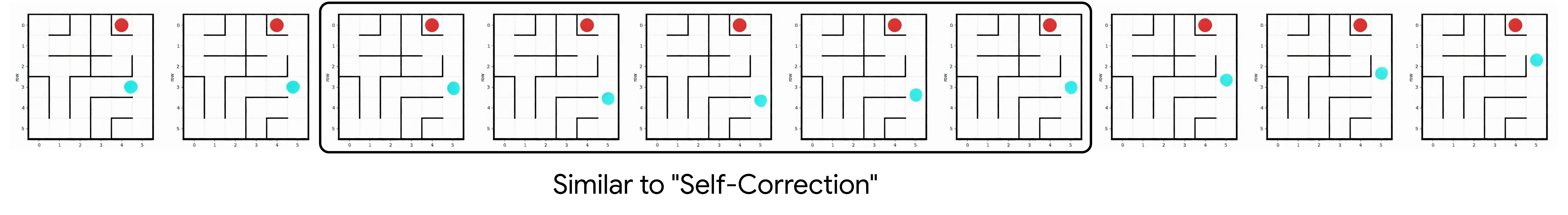}
    \caption{Example of trajectory similar to ``self-correction'' in \maze~when provided with more inference frame budget. }
    \label{appfig:maze self correction}
\end{figure*}

%% file: figures/appfig_irregular_maze.tex
\begin{figure*}[t]
    \centering
    \includegraphics[width=\linewidth]{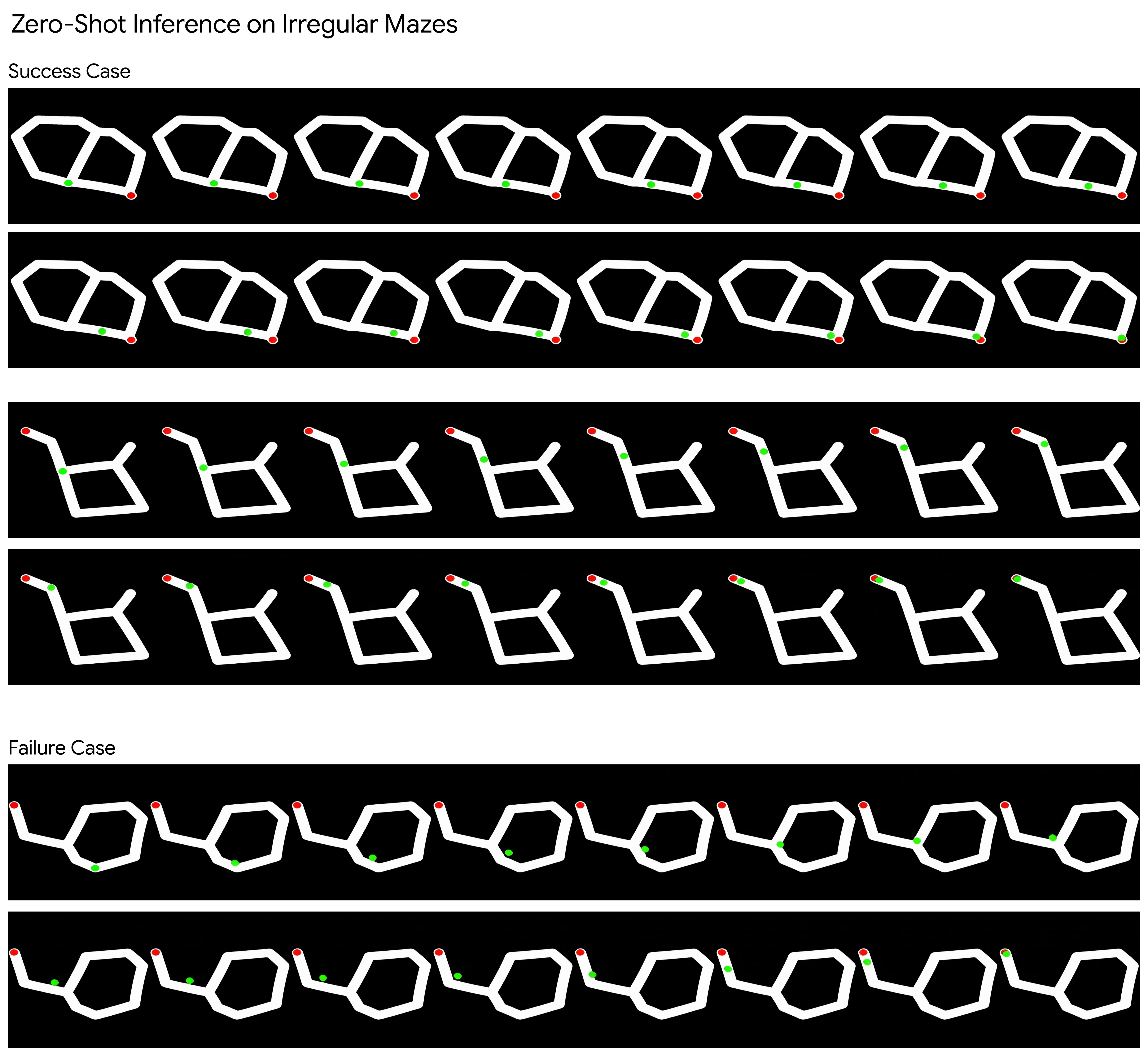}
    \caption{Visualizations of zero-shot inference on irregular maze with Wan 2.2 TI2V 5B trained on regular mazes. We observe that although not included in the training data, trained video generation model can adapt a certain level of planning capabilities to such irregular mazes with different background and meanwhile keeping the constraint (no change in the background, follow the path, do not cross the wall), and surprisingly can move diagonally. }
    \label{appfig:irregular maze}
\end{figure*}

%% file: figures/appfig_tangram_scaling.tex
\begin{figure*}[t]
    \centering
    \includegraphics[width=0.4\linewidth]{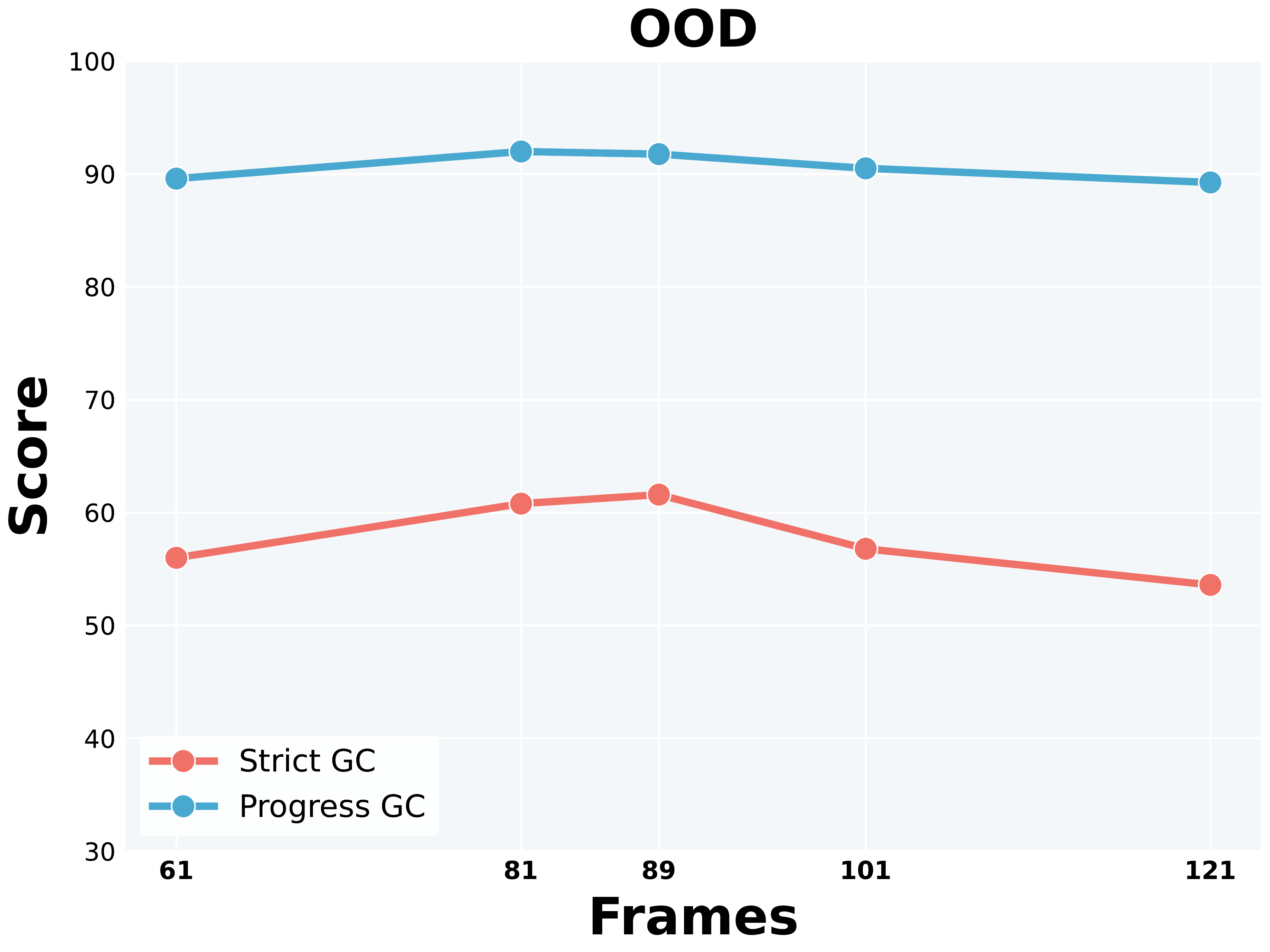}
    \caption{We don't observe similar Visual Test-Time Scaling trend in \tangram, possibly due to the constraint of geometric consistency throughout the generated video that hinders the performance. However, we still observe that by providing more inference frames (81-121 frames) compared to training setting (61-81 frames) would not severely harm the performance. }
    \label{appfig:tangram scaling}
\end{figure*}

%% file: figures/appfig_maze_correct_showcase.tex
\begin{figure*}[t]
    \centering
    \includegraphics[width=\linewidth]{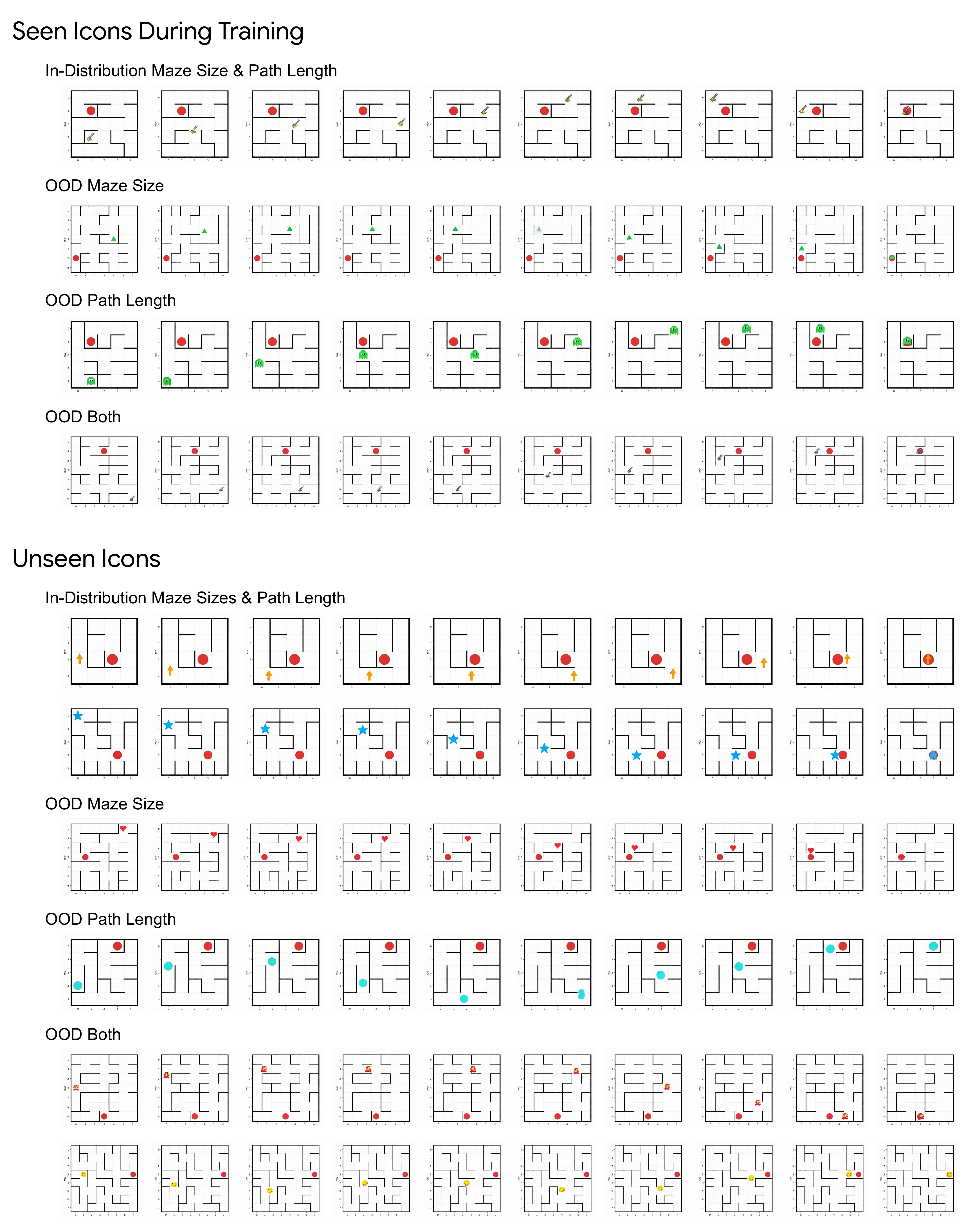}
    \caption{Correct examples of \maze~generated by fine-tuned Wan 2.2 TI2V 5B. }
    \label{appfig:maze correct showcase}
\end{figure*}

%% file: figures/appfig_maze_error_showcase.tex
\begin{figure*}[t]
    \centering
    \includegraphics[width=\linewidth]{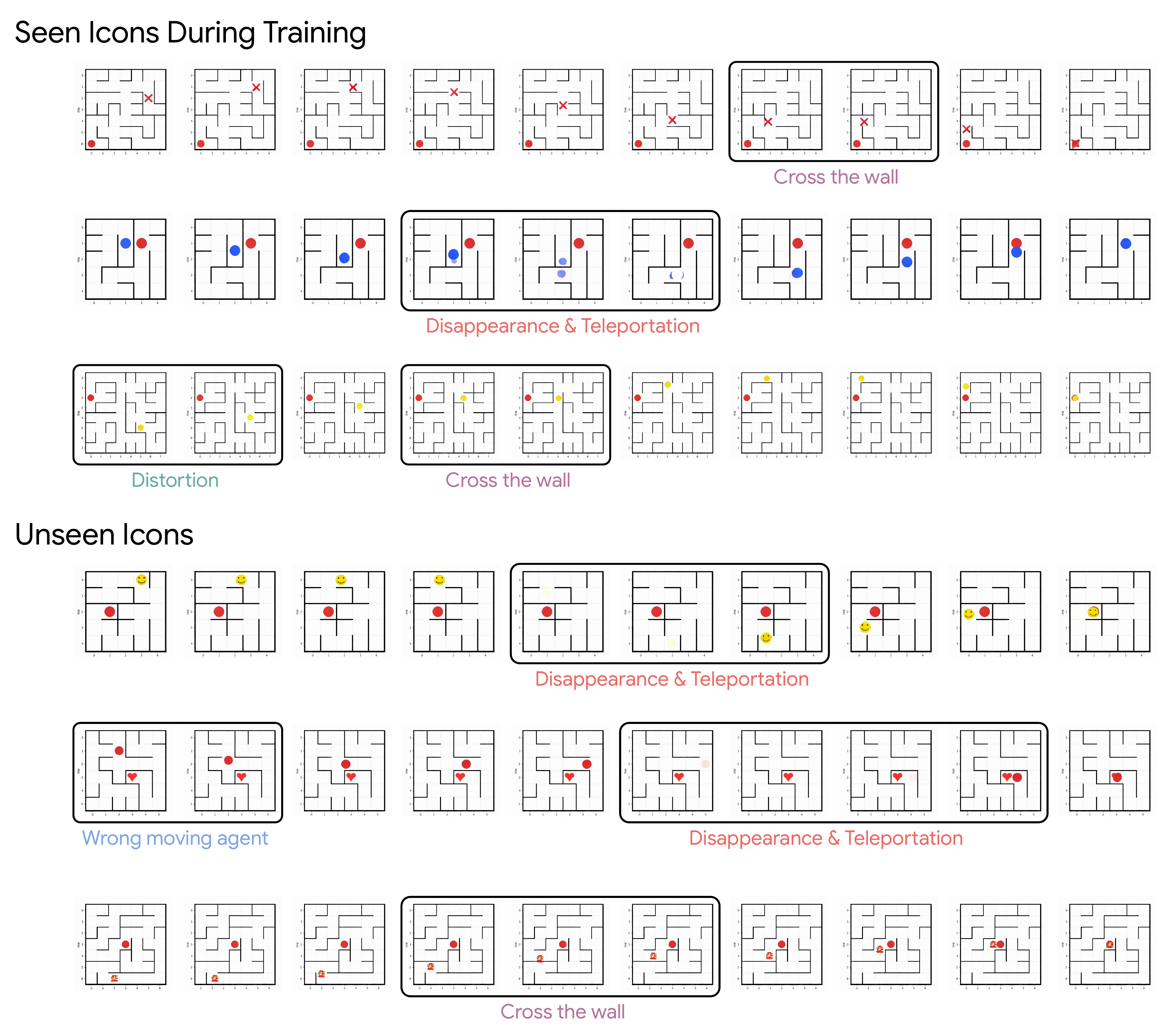}
    \caption{Wrong examples of \maze~generated by fine-tuned Wan 2.2 TI2V 5B.}
    \label{appfig:maze wrong showcase}
\end{figure*}

%% file: figures/appfig_tangram_showcase.tex
\begin{figure*}[t]
    \centering
    \includegraphics[width=\linewidth]{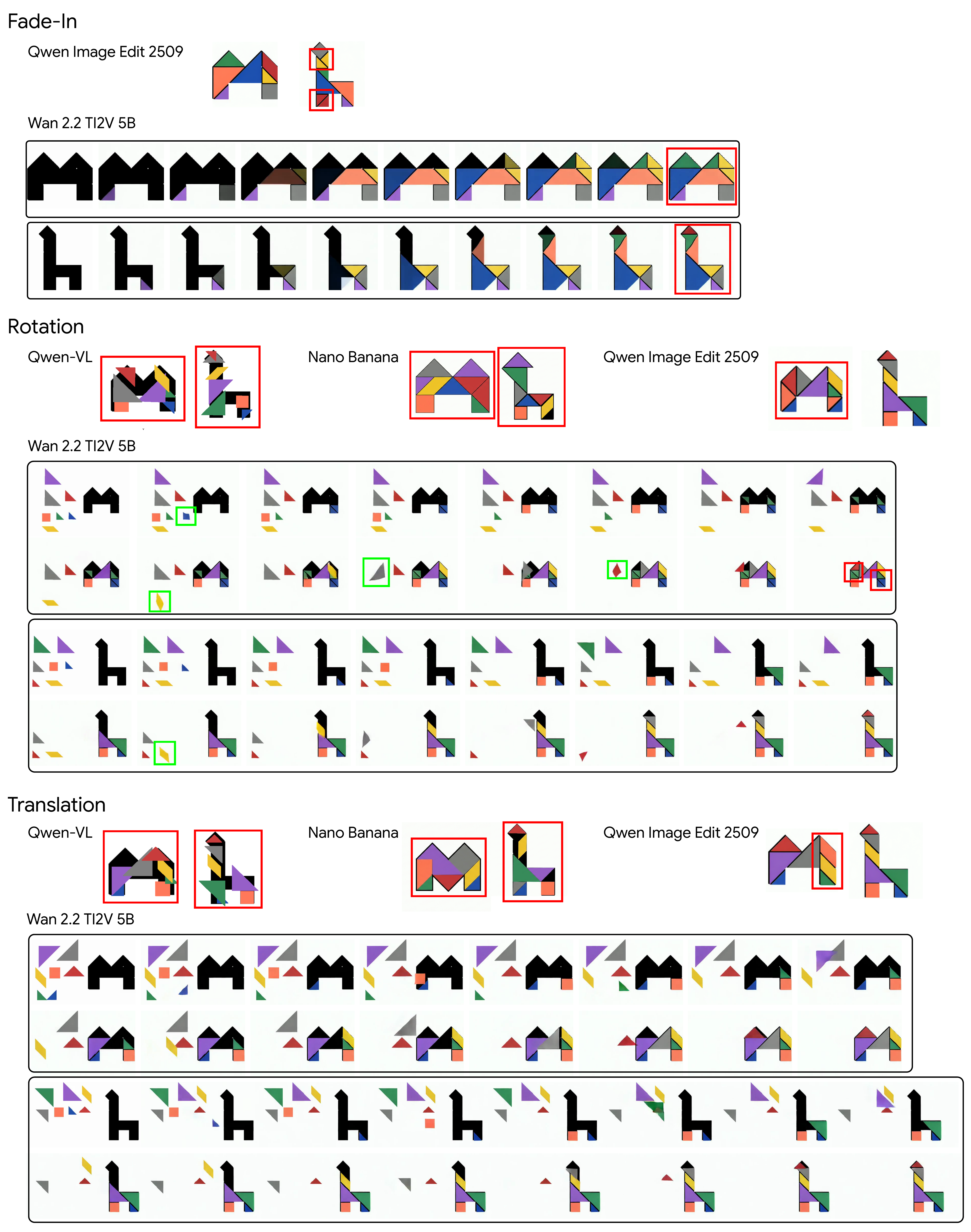}
    \caption{Examples of \tangram~generated by different systems. Red bounding boxes indicates wrong predictions, and green bounding box indicates distortion throughout the process.}
    \label{appfig:tangram showcase}
\end{figure*}

%% file: appendix/C_prompts.tex
\section{Prompt}
\label{appsec:prompt}


\begin{tcolorbox}[breakable, width=\textwidth, title=Prompt of Maze Task (Text Version)]
\footnotesize
Task : Maze Shortest Path Planning \\
\\
You are given an image of a maze environment. In this environment :\\
- An icon marks the starting position of the agent.\\
- A red point marks the goal.\\
- The agent can move in one of four cardinal directions only : "up" , "down" , "left" , or "right". Each move shifts the agent by exactly one cell in that direction. Diagonal movement is not permitted.\\
- The black maze walls are impassable. The agent cannot pass through any wall segment.\\
Your task is to analyse the image and produce the shortest valid sequence of actions that moves the agent from its starting position to the goal without crossing any wall.\\
Provide your final answer in the end enclosed between \texttt{<ANSWER>} and \texttt{</ANSWER>}, for example: \texttt{<ANSWER> right up up </ANSWER>}.
\end{tcolorbox}

\begin{tcolorbox}[breakable, width=\textwidth, title=Prompt of Maze Task (Video Version)]
\footnotesize
Create a 2D animation based on the provided image of a maze.     \\
The custom character slides smoothly along the white path, stopping perfectly on the red circle destination.  \\   
The character never slides or crosses into the black segments of the maze.     \\
The camera is a static, top-down view showing the entire maze.     \\
Maze:     \\
* The maze paths are white, the walls are black.     \\
* The character starts from its initial position.    \\ 
* The character slides smoothly along the white path.     \\
* The character never slides or crosses into the black segments of the maze.  \\
* The character stops perfectly on the red circle.     \\
Scene:     \\
* No change in scene composition.     \\
* No change in the layout of the maze.     \\
* The character travels along the path without speeding up or slowing down.     \\
Camera:     \\
* Static camera.     \\
* No zoom.     \\
* No pan.     \\
* No glitches, noise, or artifacts.\\
\end{tcolorbox}

\begin{tcolorbox}[breakable, width=\textwidth, title=Prompt of Tangram Task (Text Version)]
\footnotesize
Analyze the provided Tangram puzzle image in which the silhouette can be filled by the seven standard Tangram pieces.\\
Puzzle:\\
* The silhouette of the puzzle is represented as black area in the white background.\\
* The Tangram pieces include: 2 big triangles, 1 medium triangle, 2 small triangles, 1 square, and 1 parallelogram.\\
* The shapes of the pieces can not be altered.\\
Colors:\\
* Every piece has a distinct, unique color, as shown on the left side.\\
Question:\\
Identify the final center coordinates for each piece after it is positioned within the silhouette.\\
\end{tcolorbox}

\begin{tcolorbox}[breakable, width=\textwidth, title=Prompt of Tangram Task (Image Editing Version)]
\textbf{\texttt{Fade-In}} \\
\\
Solve the tangram puzzle by arranging the pieces to exactly match the black silhouette. Do not change the color or shape of any tangram piece, and do not modify the overall silhouette. All pieces must be placed entirely within the silhouette, without overlapping or extending beyond its boundaries.\\
\\
The colors of the tangram pieces are as follows:\\
* Large triangles: blue and orange\\
* Medium triangle: green\\
* Small triangles: purple and yellow\\
* Square: gray\\
* Parallelogram: red\\
\\
\textbf{\texttt{Rotation \& Translation}} \\
\\
Use the tangram pieces shown on the left to exactly fill the black tangram silhouette on the right. Do not change the color, size, or shape of any tangram piece, and do not alter the overall silhouette of the puzzle. All pieces must be placed entirely within the silhouette, without overlapping each other or extending beyond its boundaries.
\end{tcolorbox}

\begin{tcolorbox}[breakable, width=\textwidth, title=Prompt of Tangram Task (Video Version)]
\textbf{\texttt{Fade-In}} \\
\\
Create a 2D animation showing the step-by-step assembly of a Tangram puzzle. \\
Puzzle: \\
* The silhouette of the puzzle is represented as black area in the white background. \\
* The Tangram pieces include: 2 big triangles, 1 medium triangle, 2 small triangles, 1 square, and 1 parallelogram. \\
* The pieces appear one by one, fading in from transparent to their specific colors to fill the silhouette. \\
Colors: \\
* Big triangles: blue and orange\\
* Small triangles: purple and yellow\\
* Medium triangle: green\\
* Square: grey\\
* Parallelogram: red\\
Scene: \\
* No change in scene composition. \\
* No change in the silhouette of the puzzle. \\
Camera: \\
* Static camera. \\
* No zoom. \\
* No pan. \\
* No glitches, noise, or artifacts.\\
\\
\\
\textbf{\texttt{Rotation}}\\
\\
Create a 2D animation showing the step-by-step assembly of a Tangram puzzle. \\
Puzzle: \\
* The silhouette of the puzzle is represented as black area in the white background. \\
* The Tangram pieces include: 2 big triangles, 1 medium triangle, 2 small triangles, 1 square, and 1 parallelogram. \\
* The shapes of the pieces can not be altered. \\
* Sequential accumulation constraint: Pieces move and orient individually one by one. Upon placement, every piece is permanently locked in place with its unique color and orientation. Continue until the silhouette is full. \\
Colors: \\
* Every piece has a distinct, unique color, as shown on the left side. \\
Scene: \\
* No change in scene composition. \\
* No change in the silhouette of the puzzle. \\
* No change in the designated colors and shapes of the pieces. \\
Camera: \\
* Static camera. \\
* No zoom. \\
* No pan. \\
* No glitches, noise, or artifacts.\\
\\
\\
\textbf{\texttt{Translation}} \\
\\
Create a 2D animation showing the step-by-step assembly of a Tangram puzzle. \\
Puzzle: \\
* The silhouette of the puzzle is represented as black area in the white background. \\
* The Tangram pieces include: 2 big triangles, 1 medium triangle, 2 small triangles, 1 square, and 1 parallelogram. \\
* The shapes of the pieces can not be altered. \\
* Sequential accumulation constraint: Pieces move individually one by one. Upon placement, every piece is permanently locked in place with its unique color and orientation. Continue until the silhouette is full. \\
Colors: \\
* Every piece has a distinct, unique color, as shown on the left side. \\
Scene: \\
* No change in scene composition. \\
* No change in the silhouette of the puzzle. \\
* No change in the designated colors and shapes of the pieces. \\
Camera: \\
* Static camera. \\
* No zoom. \\
* No pan. \\
* No glitches, noise, or artifacts.\\
\end{tcolorbox}